\documentclass[11pt]{article}

\usepackage[final]{acl}

\usepackage{times}
\usepackage{latexsym}

\usepackage[T1]{fontenc}
\usepackage[utf8]{inputenc}

\usepackage{microtype}

\usepackage{inconsolata}

\usepackage{graphicx}
\usepackage{hyperref}       % hyperlinks
\usepackage{url}            % simple URL typesetting
\usepackage{booktabs}       % professional-quality tables
\usepackage{amsfonts}       % blackboard math symbols
\usepackage{nicefrac}       % compact symbols for 1/2, etc.
\usepackage{xcolor}         % colors
\usepackage{amsmath}
\usepackage{amssymb}
\usepackage{xspace}
\usepackage{multirow}
\usepackage{wrapfig}
\usepackage{float}
\usepackage{enumitem}
\usepackage{caption}
\usepackage{subcaption}
\usepackage[most]{tcolorbox}
\usepackage{CJK}

\tcbuselibrary{skins, breakable}

\title{Scaling External Knowledge Input Beyond Context Windows of LLMs \\via Multi-Agent Collaboration}

\author{Zijun Liu\thanks{\quad Equal contribution.}\textsuperscript{1}, Zhennan Wan\protect\footnotemark[1]\textsuperscript{1}, Peng Li\thanks{\quad Corresponding authors.}\textsuperscript{2}, Ming Yan\protect\footnotemark[2]\textsuperscript{3}, Fei Huang\textsuperscript{3}, Yang Liu\protect\footnotemark[2]\textsuperscript{1,2} \\
  \textsuperscript{1}Dept. of Comp. Sci. \& Tech., Institute for AI, Tsinghua University, Beijing, China \\
  \textsuperscript{2}Institute for AI Industry Research (AIR), Tsinghua University, Beijing, China \\
  \textsuperscript{3}Institute of Intelligent Computing, Alibaba Group \\
  \texttt{zj-liu24@mails.tsinghua.edu.cn}, \texttt{zhennan018@gmail.com}, \\ \texttt{pengli09@gmail.com}, \texttt{ym119608@alibaba-inc.com}, \texttt{liuyang2011@tsinghua.edu.cn} \\}

\begin{document}

\begin{CJK*}{UTF8}{gbsn}

\newcommand{\ExtAgents}{\textsc{ExtAgents}\xspace}

\maketitle

% \renewcommand{\thefootnote}{\fnsymbol{footnote}}
% \footnotetext[1]{Equal contribution.}
% \footnotetext[2]{Code and data are available at \url{https://github.com/THUNLP-MT/ExtAgents}.}

\begin{abstract}
With the rapid advancement of post-training techniques for reasoning and information seeking, large language models (LLMs) can incorporate a large quantity of retrieved knowledge to solve complex tasks. %In general, more knowledge input leads to better performance. 
However, the limited context window of LLMs obstructs scaling the amount of external knowledge input, prohibiting further improvement. % especially for tasks requiring significant amount of external knowledge. 
Existing context window extension methods inevitably cause information loss. 
LLM-based multi-agent methods emerge as a new paradigm to handle massive input in a distributional manner, where we identify two core bottlenecks in existing agent orchestration designs. 
In this work, we develop a multi-agent framework, \textbf{\ExtAgents}, to overcome the bottlenecks and enable better scalability in inference-time knowledge integration without longer-context training. 
Benchmarked with our enhanced multi-hop question answering test, \textbf{$\boldsymbol{\infty}$Bench+}, and other public test sets including long survey generation, 
\ExtAgents significantly enhances the performance over existing non-training methods with the same amount of external knowledge input, regardless of whether it falls \emph{within or exceeds the context window}. Moreover, the method maintains efficiency due to high parallelism. We believe further study in the coordination of LLM agents on increasing external knowledge input could benefit real-world applications.\footnote{\quad Code and data are available at \url{https://github.com/THUNLP-MT/ExtAgents}.} 
\end{abstract}

\section{Introduction}
\label{sec:intro}

% \begin{wrapfigure}[15]{r}{0.43\linewidth}
%   \centering
%   \vspace{-5em}
%   \includegraphics[width=0.9\linewidth]{./figs/fig-trend-all.pdf}
%   \caption{Performance of scaling external knowledge input with \ExtAgents and LLM$\times$MapReduce~\citep{zhou2024llmtimesmapreducesimplifiedlongsequenceprocessing} on $\infty$Bench+.}
%   \label{fig:top-trend}
% \end{wrapfigure}

% \looseness=-1 
Large Language Models (LLMs) have recently witnessed dramatic progress in parameter scales and context lengths, culminating in context windows that span more than a book-length of text~\citep{deepseekai2025deepseekv3technicalreport,openai2025gpt5,anthropic2025claude45sonnet}. 
%\footnote{For instance, GPT-4o‐128k and Claude‐3‐200k can already process summaries of whole books.}  
Yet even these impressive limits remain insufficient for many real-world tasks---multi-hop question answering with Internet, reasoning over enterprise knowledge bases, or writing surveys based on massive academic research---where more external knowledge input often results in better outcomes. 
Especially, recent research on post-training LLMs to generate long chains of thoughts on reasoning~\citep{openai2025o3o4mini,deepseekai2025deepseekr1incentivizingreasoningcapability} and information seeking~\citep{openai2025deepresearch,li2025searcho1agenticsearchenhancedlarge,song2025r1searcherincentivizingsearchcapability,jin2025searchr1trainingllmsreason} tasks, has shown that increasing the amount of retrieved knowledge within the context window could lead to better task performance~\citep{yue2025inferencescalinglongcontextretrieval}.  

\begin{figure}[t]
  \centering
  \fbox{\includegraphics[width=0.67\linewidth]{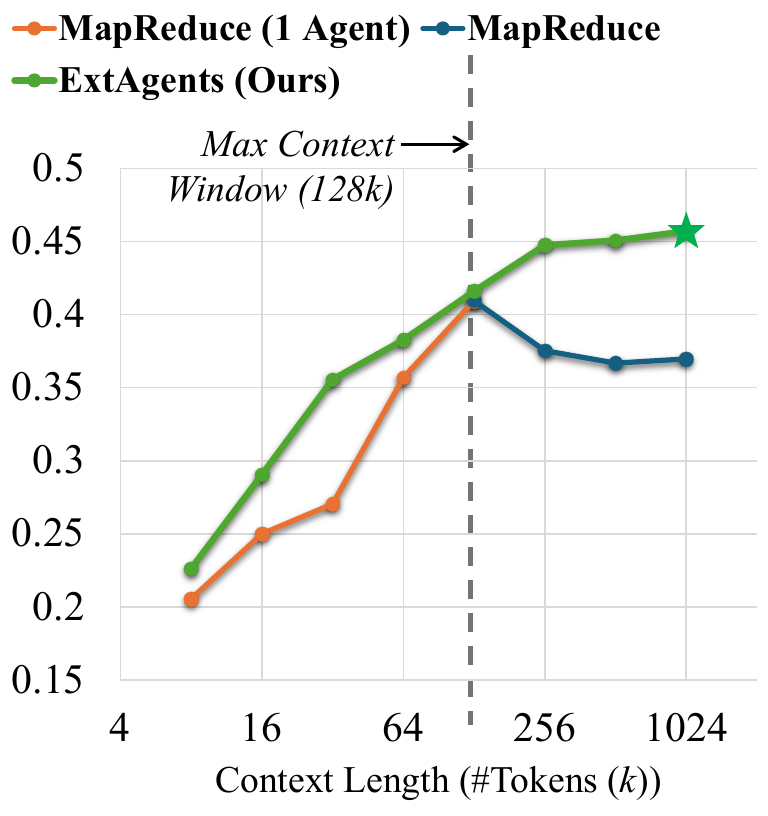}}
  \caption{Performance of scaling external knowledge input with \ExtAgents (Ours) and LLM$\times$MapReduce \citep{zhou2024llmtimesmapreducesimplifiedlongsequenceprocessing} on $\infty$Bench+ (detailed in Figure~\ref{fig:scaling-exp}).}
  \label{fig:top-trend}
  % \vspace{-1.5em}
\end{figure}

% \looseness=-1 
For larger input beyond the context window length, the situation is more complicated. 
When such knowledge is crudely truncated, essential evidence is lost and downstream performance suffers. 
A natural solution is to train ever longer-context models~\citep{chen2023extendingcontextwindowlarge,peng2023yarnefficientcontextwindow,xu2025128k4mefficienttraining,shang2025longrope2nearlosslessllmcontext}, but this is economically prohibitive and practically brittle: (i) the quadratic complexity of attention~\citep{10.5555/3295222.3295349} becomes intractable; and (ii) longer-context training data is scarce. 
Consequently, practitioners turn to \emph{retrieval-augmented generation} (RAG)~\citep{lewis2021retrievalaugmentedgenerationknowledgeintensivenlp,gao2024retrievalaugmentedgenerationlargelanguage,packer2024memgptllmsoperatingsystems} or \emph{context compression}~\citep{jiang-etal-2024-longllmlingua,qian2024longllmsnecessitylongcontexttasks,xiao2024efficient,wang2024with,hao2025omnikv} pipelines. Unfortunately, both strategies inevitably introduce information loss: RAG is limited by ranking errors that could exclude essential evidence during retrieval, while compressors may discard subtle cues that are only useful once the reasoning chain unfolds. 
Recent approaches~\citep{trivedi-etal-2023-interleaving,zhao-etal-2024-longagent,zhang2024chain,zhou2024llmtimesmapreducesimplifiedlongsequenceprocessing} let LLM-based agents collaborate to process long contexts distributedly, reaching state-of-the-art performance on long-context tasks. 
In this work, we take a step further by asking a question: \textbf{Could LLMs consistently improve task performance by scaling the amount of external knowledge input beyond the context window?}
Achieving high scalability of external knowledge implies two requirements: (i) a scalable context extension method needs to accept the massive input, and (ii) the knowledge should be effectively integrated in the orchestration of LLMs and agents. 
Since it is impractical to re-train short-context LLMs, we mainly focus on the scalability of \emph{inference-time knowledge integration beyond context windows}. 

We focus on a few tasks that require massive external knowledge, including multi-hop question answering (QA), both over long documents and large knowledge bases, and long survey generation. % (Figure~\ref{fig:scaling}). 
We found current benchmarks on long-context tasks constructed with biases, that a quantity of queries could be answered by sweeping a small context window over the attached document. 
For comprehensive validation, we enhance the existing long-context benchmark, $\infty$Bench~\citep{zhang-etal-2024-bench}, with an automated pipeline, to obtain a long-document-based multi-hop QA evaluation set, \textbf{$\boldsymbol{\infty}$Bench+}, alongside with public multi-hop QA~\citep{yang-etal-2018-hotpotqa} and long survey generation~\citep{wang2024autosurvey} benchmarks. 
% We demonstrate the effectiveness and efficiency of \ExtAgents with comprehensive experiments on the aforementioned benchmarks. 

% \begin{figure}[t]
%   \centering
%   \includegraphics[width=0.85\linewidth]{./figs/fig-scaling.pdf}
%   \caption{The illustration of scaling external knowledge input for context window extension methods for LLMs. Ideally, knowledge-intensive tasks, including QA and long generation, should benefit from scaled input.}
%   \label{fig:scaling}
%   \vspace{-1.5em}
% \end{figure}

\looseness=-1 
In preliminary experiments, we find that the current state-of-the-art LLM-based multi-agent system~\citep{zhou2024llmtimesmapreducesimplifiedlongsequenceprocessing} fails to consistently improve task performance with scaled external knowledge input, and even degrades the performance compared to truncated input (Figure~\ref{fig:top-trend}). 
We systematically analyzed existing multi-agent methods, and 
then spotted two core bottlenecks in the shared designs of agent orchestration: (i) \emph{knowledge synchronization} that agents comprehend the distributed contexts and provide condensed information, where the bottleneck is the ``bandwidth'' of accessible agents for each agent. 
and (ii) \emph{knowledge-integrated reasoning}, where the bottleneck is the ratio of redundant information in the reasoning process. 
To overcome the bottlenecks, we develop a scalable multi-agent framework, \textbf{\ExtAgents}. 
Following prior distributional paradigm, the framework partitions the full input into agent-specific context chunks, each sized to fit a small window. 
\ExtAgents simplifies the roles of agents into two: Seeking Agents and Reasoning Agent; featuring two key components: \emph{global knowledge synchronization}, where Seeking Agents to globally exchange and update salient intermediate results instead of locally sharing entire context chunks~\citep{zhao-etal-2024-longagent,zhou2024llmtimesmapreducesimplifiedlongsequenceprocessing}, and \emph{knowledge-accumulating reasoning}, which gradually integrates and increases the updated knowledge from Seeking Agents to Reasoning Agent throughout multiple rounds of reasoning.

We demonstrate the effectiveness and efficiency of \ExtAgents with comprehensive experiments on the aforementioned benchmarks. 
We show that \ExtAgents consistently improves task performance with scaled external knowledge input, outperforming the state-of-the-art non-training methods and achieves increasing performance when the input exceeds context windows. We show the generalization of \ExtAgents across different QA and long generation tasks, and its compatibility with different LLM families. 
We also measure the efficiency gain of \ExtAgents from high parallelism. 

In summary, our contributions are:
\begin{itemize} %[itemsep=0em, topsep=0em]
    \item We \emph{introduce and define} the problem of \textbf{scaling external knowledge input beyond context windows}, filling a critical gap in current LLM deployment. We also construct an enhanced multi-hop QA benchmark, \textbf{$\boldsymbol{\infty}$Bench+}, for corresponding evaluation.  
    \item We systematically study existing LLM-based multi-agent systems for context window extension, and overcome their bottlenecks by proposing a novel framework, \ExtAgents.  
    \item We demonstrate the effectiveness and efficiency of \ExtAgents on QA and survey generation tasks. With external knowledge input scaling beyond context windows, it consistently improves task performance and significantly outperforms baseline methods. 
\end{itemize}

\section{Related Work}
\label{sec:related}
% \vspace{-1em}

\paragraph{Context Window Extension Methods for LLMs}
%We group existing approaches into three main categories: 
(1) \textit{\textbf{Retrieval-Based Methods:}} % Token-Level Retrieval
    For massive input breaking the context window, RAG~\citep{lewis2021retrievalaugmentedgenerationknowledgeintensivenlp,gao2024retrievalaugmentedgenerationlargelanguage} is a common solution to chunk the input into smaller pieces and retrieve relevant ones through indexing~\citep{10.1145/3637870}, searching~\citep{packer2024memgptllmsoperatingsystems}, or ranking~\citep{wang2024xl3mtrainingfreeframeworkllm}. The granularity ranges from token-~\citep{xiao2024infllm} to document-level~\citep{chen2025ultraragmodularautomatedtoolkit}. 
    Recently, iterative retrieval is shown to be effective for multi-hop tasks~\citep{trivedi-etal-2023-interleaving} and with scaled retrieved documents~\citep{yue2025inferencescalinglongcontextretrieval}. Since the amount of acceptable retrieved information is limited by the context window, ranking errors are the decisive factor in performance. 
    (2) \textit{\textbf{Compression-Based Methods:}} % Parametric Memory
    Orthogonal to RAG, long contexts can be compressed into smaller representations, including parametric states~\citep{han-etal-2024-lm,xiao2024efficient,wang2024with,hao2025omnikv,yang-etal-2025-rethinking} and non-parametric summaries~\citep{chen2023walkingmemorymazecontext,jiang-etal-2024-longllmlingua,qian2024longllmsnecessitylongcontexttasks,edge2025localglobalgraphrag}, which are then fed into LLMs. However, the compression is often lossy due to the limited context window and compression capabilities of compressor model. 
    (3) \textit{\textbf{Multi-Agent Collaboration Methods:}}
    LLM-based multi-agent systems have emerged as a new paradigm to handle massive input in a distributional manner. We analyze existing methods~\citep{zhao-etal-2024-longagent,zhang2024chain,zhou2024llmtimesmapreducesimplifiedlongsequenceprocessing, li-etal-2024-graphreader, wang2025llmtimesmapreducev2entropydrivenconvolutionaltesttime} in Section~\ref{sec:review} in detail. 
    Though the approach could be viewed as mixing retrieval and compression, it contains more nuanced orchestration of agents. 

\paragraph{LLM-Based Multi-Agent Collaboration on General Tasks}
For general tasks, LLM-based multi-agent systems~\citep{li2023camel,pmlr-v235-du24e,wu2024autogen,liu2024a,pmlr-v235-zhuge24a,zhang2025gdesignerarchitectingmultiagentcommunication,yue2025masrouterlearningroutellms} have been proposed to collaboratively process workloads, resulting in improved performance. Different applications has been explored, including coding~\citep{hong2024metagpt}, science research~\citep{yamada2025aiscientistv2workshoplevelautomated}, decision making~\citep{wang2024mobileagentv}, embodied game playing~\citep{chen2024agentverse}, etc. Recent studies also post-train LLMs to enhance collaboration in various tasks~\citep{qiao-etal-2024-autoact,subramaniam2025multiagent,he2025enhancinglanguagemultiagentlearning,liao2025marftmultiagentreinforcementfinetuning}.

% \vspace{-0.5em}
\section{Scaling External Knowledge Input Beyond Context Windows of LLMs}

\begin{figure}[t]
  \centering
  \includegraphics[width=0.95\linewidth]{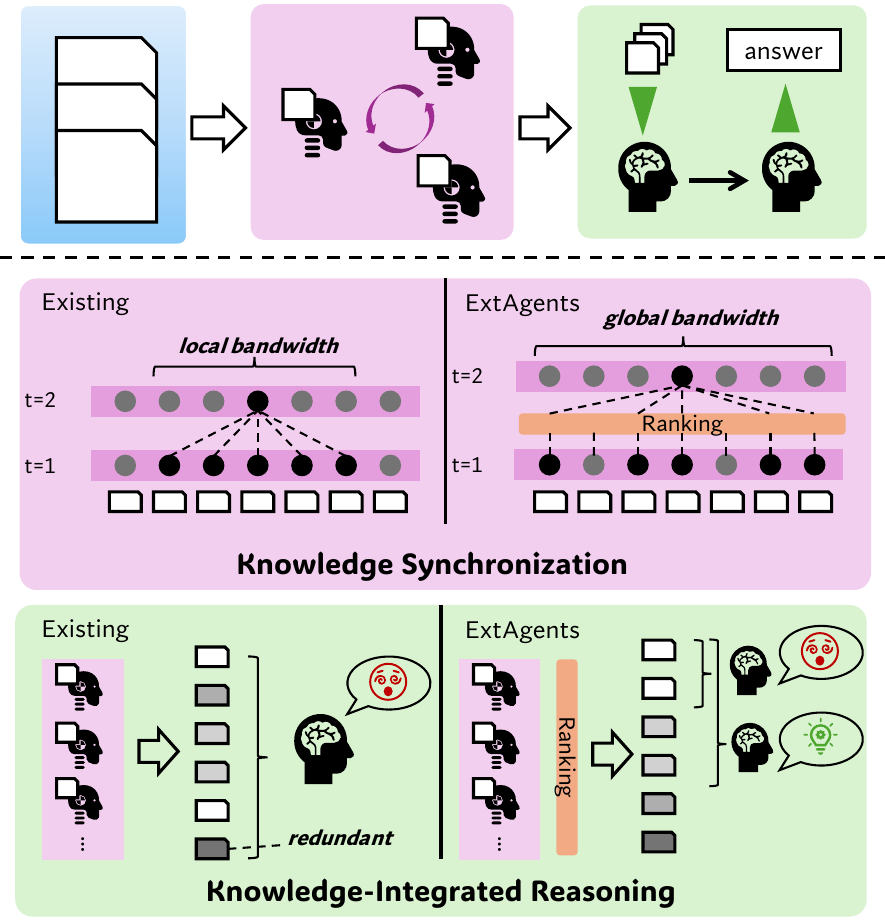}
  \caption{The illustration of multi-agent collaboration methods for context window extension on LLMs. \ExtAgents alleviates bottlenecks in knowledge synchronization and reasoning processes (detailed in Table~\ref{tab:existing-methods}).}
  \label{fig:scaling}
  % \vspace{-1.5em}
\end{figure}

\begin{table*}[t]
    \centering
    \resizebox{0.85\linewidth}{!}{
    \begin{tabular}{lccc}
    \toprule
    \textbf{Method} & \textbf{Sync.~Bandwidth} &  \textbf{Reasoning Context} ($\mathcal{M}_r$) & \textbf{Parallelized Component}\\
    \midrule
    Chain of Agents~\citep{zhang2024chain} &
    $2$ &
    $\{m_{N,N}\}$ &
    None\\
    LongAgent~\citep{zhao-etal-2024-longagent} &
    $2$ &
    $\{m_{i,t}\}_{1\le i\le N, 1\le t\le T}$ &
    Sync.\\
    LLM$\times$MapReduce~\citep{zhou2024llmtimesmapreducesimplifiedlongsequenceprocessing} &
    $O(\frac{L}{\lvert m\rvert})$ &
    $\{m_{i,T}\}_{1\le i\le N}$ &
    Sync.\\ \midrule
    \textbf{ExtAgents} (Ours) &
    $\boldsymbol{N}$ &
    $\boldsymbol{\mathrm{Top}_{2^s} (\{m_{i,t^\star}\}_{1\le i\le N})}$ &
    \textbf{Sync.~\& Reasoning}\\
    \bottomrule
    \end{tabular}
    }
    \caption{Comparisons of existing LLM-based multi-agent methods for context window extension and our \ExtAgents on the bottlenecks in agent orchestration of knowledge synchronization (sync.) and reasoning processes.}
    \label{tab:existing-methods}
    % \vspace{-1.5em}
\end{table*}

\vspace{-0.5em}
\subsection{Problem Definition}
\label{sec:problem}

% TODO: different to scaling retrieval knowledge base

Since real-world applications~\citep{openai2025deepresearch,wei2025browsecompsimplechallengingbenchmark,deerflow2025} frequently demand extensive external knowledge whose scale could dramatically surpass LLM context windows, the need for a scalable approach is paramount. % (Figure~\ref{fig:scaling}). 

\noindent
\textbf{Formal View\quad} 
For each task query $q\in\mathcal{Q}$, a given external knowledge source $\mathcal{K}$ could be a long document attached ($\mathcal{K} = f_{\mathcal{Q}}(q)$) or several document pieces retrieved from large knowledge bases $\mathcal{C}$ ($\mathcal{K} = f_{\mathcal{C}}(q)$). 
In both cases, the knowledge source could be partitioned into $N$ chunks $\mathcal{K} = \{d_1,\dots,d_N\}$, where $d_i$ is a chunk with length $\lvert d_i\rvert$. For long documents, this could be done by simple splitting, and sophisticated chunking methods are available for future work; for knowledge bases $\mathcal{C}$, chunk $d_i$ could be a retrieved document piece with further aggregation or splitting. 
The former is often used in QA tasks attached to long documents~\citep{zhang-etal-2024-bench}, while the latter is common in open-domain tasks~\citep{yang-etal-2018-hotpotqa}. 
The query is processed by an LLM $\theta$ with a maximum context length $L,  \max_i \{\lvert d_i\rvert\} < L$ (e.g., 128k, or more tokens), under the guidance of pre-defined prompts and agent orchestration (workflows) $\pi_\theta$, to give out answer $y$. 
In this work, we focus on tasks where the total length of knowledge source is much larger than the context window, i.e., %$\lvert \mathcal{K}\rvert \gg L$: 
\vspace{-0.5em}
\begin{equation}
    y = \pi_\theta\bigl(q, \mathcal{K}\bigr) \quad\text{with}\quad \lvert\mathcal{K}\rvert \gg L.
% \vspace{-0.5em}
\end{equation}

\vspace{-0.4em}
\noindent
\textbf{Objective\quad}  
The overall objective is to maximize the task performance with respect to the amount of external knowledge input. 
For tasks with ground-truth answers $y^\star$, the objective is formulated as:
%\vspace{-1em}
\begin{equation}
\resizebox{0.85\columnwidth}{!}{$
\begin{aligned}
    \max_{\pi}\;
        \mathbb{E}_{q\sim\mathcal{Q}, \mathcal{K}\sim \{f_{\mathcal{Q}}(q), f_{\mathcal{C}}(q)\}}\bigl[\mathrm{Score}_{\mathrm{pair}}\bigl(y,\,y^\star\bigr)\bigr]
        \\
        \quad\text{with fixed}\quad \max\{\lvert \mathcal{K}\rvert\}, \theta
\end{aligned}
$}
\end{equation}
% \vspace{-0.5em}
where $\mathrm{Score}_{\mathrm{pair}}(\cdot, \cdot) \in \mathbb{R}$ is a specific metric with reference (e.g., F1, LLM-as-a-Judge~\citep{zheng2023judging}). For open-ended generation tasks without clear ground truths, the objective is 
\vspace{-0.6em}
\begin{equation}
  % \scalebox{0.93}{$
\begin{aligned}
    \max_{\pi}\;
        \mathbb{E}_{q\sim\mathcal{Q}, \mathcal{K}\sim \{f_{\mathcal{Q}}(q), f_{\mathcal{C}}(q)\}}
        \bigl[\mathrm{Score}_{\mathrm{single}}\bigl(y\bigr)\bigr]
        \\ 
        \quad\text{with fixed}\quad \max\{\lvert \mathcal{K}\rvert\}, \theta
\end{aligned}
% $}
\vspace{-0.6em}
\end{equation}
where $\mathrm{Score}_{\mathrm{single}}(\cdot) \in \mathbb{R}$ is a reference-free metric (e.g., LLM with rating principles~\citep{wang2024autosurvey}). 
The control of maximum input length $\max\{\lvert \mathcal{K}\rvert\}$ is achieved by truncating $f_{\mathcal{Q}}(\cdot)$ or $f_{\mathcal{C}}(\cdot)$. 

Noticeably, the setting of scaling external knowledge input is different from expanding retrieval knowledge bases~\citep{shao2024scaling}, which does not increase inference costs of LLMs but of the retriever. We argue this is orthogonal to our primary goal towards the scalability of LLM-based agents.

\subsection{Review of Existing Multi-Agent Methods}
\label{sec:review}

In this section, we review representative LLM-based multi-agent systems for context window extension, including \textbf{Chain of Agents}~\citep{zhang2024chain}, \textbf{LongAgent}~\citep{zhao-etal-2024-longagent}, and \textbf{LLM$\boldsymbol{\times}$MapReduce}~\citep{zhou2024llmtimesmapreducesimplifiedlongsequenceprocessing}. 
These methods spin up a team of $N$ LLM-based agents. Each agent is attributed a local context chunk $d_i$, and collectively decides on the answer $y$. 
We conclude that these multi-agent methods share a two-stage pattern of \emph{knowledge synchronization} and \emph{reasoning}. The former is designed to comprehend the distributed contexts and provide related knowledge for the latter to generate the final answer. We follow \citet{liu2024a} to incorporate timesteps for modeling agent orchestration, and we identify a core bottleneck in each stage (Figure~\ref{fig:scaling}):

\looseness=-1
\begin{enumerate}[leftmargin=1em, nosep]
    \item \textbf{Knowledge Synchronization:}  
          At timestep $t\le T$, each agent $a_{i,t}$ digests its local chunk $d_i$ and messages $\mathcal{M}_{\mathcal{G}_{i,t-1},t-1}\subseteq\{m_{j,t-1}|a_{j,t-1}\in\mathcal{G}_{i,t-1}\}$ from a neighbourhood
          $\mathcal{G}_{i,t-1} = \{a_{i-k_1,t-1},\dots,a_{i,t-1},\dots,a_{i+k_2,t-1}\}$ of size $\lvert \mathcal{G}_{i,t-1}\rvert$, with the maximum $\max_{i,t}\{\lvert\mathcal{G}_{i,t}\rvert\}$ termed \emph{bandwidth}. Original chunks may also be included ($\mathcal{D}_{\mathcal{G}_{i,t-1},t} \subseteq \{d_i|a_{i,t-1}\in\mathcal{G}_{i,t-1}\}$). 
          It is then prompted to emit an updated message:
          \vspace{-0.61em}
          \begin{equation}\label{eq:sync}
            m_{i,t} = a_{i,t} (q, \mathcal{D}_{\mathcal{G}_{i,t-1},t}, \mathcal{M}_{\mathcal{G}_{i,t-1},t-1}). 
          \end{equation}
          With smaller bandwidths, more timesteps might be needed to synchronize all the inputs. 
          The bandwidth of Chain of Agents and LongAgent is $2$, and the bandwidth of LLM$\times$MapReduce is $O\left(\frac{L}{\lvert m\rvert}\right)$, where $\lvert m\rvert$ is the expected length of a single message. The \underline{values of bandwidth} generally reflect the reported performance~\citep{zhou2024llmtimesmapreducesimplifiedlongsequenceprocessing}, and we conjecture larger bandwidth leads to better performance. 
    \item \textbf{Knowledge-Integrated Reasoning:}  
          An agent $a_r$ collects a subset of messages as the \emph{reasoning context} $\mathcal{M}_r\subseteq\{m_{i,t}\}_{i,t}$ and produces the task answer following a workflow:
          \vspace{-0.75em}
          \begin{equation}\label{eq:reason}
            y = a_r\bigl(q, \mathcal{M}_r\bigr).
          \vspace{-0.75em}
          \end{equation}
          According to \citet{jiang-etal-2024-longllmlingua}, the \underline{ratio of} \underline{redundant information} in the reasoning process is a key factor affecting the performance. Chain of Agents and LLM$\times$MapReduce default to put as much information as possible into the reasoning context, which may lead to information overload. 
\end{enumerate}
Details are shown in Table~\ref{tab:existing-methods} and Appendix~\ref{app:review-detail}. 
Other multi-agent systems with additional designs orthogonal to the agent orchestration could be analyzed similarly, e.g., \citet{li-etal-2024-graphreader} is similar to LongAgent with a graph to organize retrieved information. 
Due to the high costs of long inputs, we use LLM$\boldsymbol{\times}$MapReduce and Chain of Agents as main baselines in our experiments.

\subsection{Challenges: Evaluation and Implementation of Scalable Approaches}
\label{sec:challenge}

\paragraph{$\boldsymbol{\infty}$Bench+}
\citet{zhang-etal-2024-bench} introduced $\infty$Bench to evaluate LLMs on long-document inputs. However, many samples exhibit bias, as answer-related content is confined to small regions easily handled within limited windows. To address this, we construct \textbf{$\boldsymbol{\infty}$Bench+}, an enhanced multi-hop QA test set requiring the benchmarked system to aggregate information across large segments of each document.  Using gpt-4o-mini-2024-07-18~\citep{openai2024gpt4ocard}, we discard samples answerable with any 8k-token chunk, thereby exposing bias in some long examples (Table~\ref{tab:bench-stat}). The 8k-token chunk represents the standard that (1) the vast majority of modern LLMs can process \emph{trivially}, and (2) is the shortest chunk size in \emph{subsequent experiments} (Section~\ref{sec:experiments}). Sensitivity checks (Table~\ref{tab:bench-sensitivity}) demonstrate that 8k-token sweeping window is robust, which is less coarse compared to 16k (remaining \textasciitilde 2\% more samples) or 32k (\textasciitilde 4\%) and effectively eliminates the ``shortcut'' queries without aggressively decimating the dataset. As this filtering results in a limited number of samples, making evaluation unstable, we additionally incorporate original $\infty$Bench samples longer than 128k tokens into $\infty$Bench+. The resulting $\infty$Bench+ consists of two subsets: En.QA with 294 samples and Zh.QA with 184 samples. We claim that $\infty$Bench+ the benchmark does not significantly shift from but enhance the existing long-context QA benchmarks~\citep{hsieh2024ruler, yen2025helmet, bai-etal-2024-longbench}.

\begin{table}[t]
  \centering
  \resizebox{0.8\linewidth}{!}{
  \begin{tabular}{lccc}
    \toprule
    \textbf{Subset} & \textbf{Samples} & \textbf{\#Samples} & \textbf{\#Tokens} (Avg.) \\
    \midrule
    \multirow{2}{*}{En.QA} & All & 351  & \textasciitilde194k  \\
    & $\lnot$8k & \textbf{157}  &  \textbf{\textasciitilde188k}  \\ \midrule
    \multirow{2}{*}{Zh.QA} & All & 189  & \textasciitilde 1,302k \\
     & $\lnot$8k & \textbf{56}  & \textbf    {\textasciitilde904k}  \\
    \bottomrule
  \end{tabular}
  }
  %\caption{Statistical information of the En.QA and Zh.QA subsets in $\infty$Bench~\citep{zhang-etal-2024-bench}. ``$\lnot$8k'' denotes samples answerable with a 8k-token chunk are filtered out.}
  \caption{Statistical information of $\infty$Bench~\citep{zhang-etal-2024-bench}. ``$\lnot$8k'' denotes samples answerable with an 8k-token chunk are {\it filtered out}.}
  \label{tab:bench-stat}
  % \vspace{-1.5em}
\end{table}

\begin{table}[t]
  \centering
  \resizebox{0.65\linewidth}{!}{
  \begin{tabular}{lccc}
    \toprule
    \textbf{Subset} & \textbf{Samples} & \textbf{\#Samples} \\
    \midrule
    \multirow{5}{*}{En.QA} & All ($\infty$Bench) & 351    \\ \cmidrule{2-3}
    & $\lnot$4k & 145    \\ 
    & \textbf{$\lnot$8k ($\infty$Bench+)}  & \textbf{157}   \\ 
    & $\lnot$16k & 164  \\ 
    & $\lnot$32k & 171  \\ 
    \bottomrule
  \end{tabular}
  }
  %\caption{Statistical information of the En.QA and Zh.QA subsets in $\infty$Bench~\citep{zhang-etal-2024-bench}. ``$\lnot$8k'' denotes samples answerable with a 8k-token chunk are filtered out.}
  \caption{Sensitivity checks on different sweeping window lengths on the En.QA subset of $\infty$Bench. The numbers of remaining valid samples are reported.}
  \label{tab:bench-sensitivity}
  % \vspace{-1.5em}
\end{table}

\paragraph{Preliminary Experiments}
We test LLM$\times$Map-Reduce on our $\infty$Bench+ benchmark with gpt-4o-mini-2024-07-18, and find that the method fails to consistently improve task performance with gradually increasing external knowledge input from 8k tokens (Figure~\ref{fig:top-trend}). When scaling the input beyond the context window of 128k tokens, the performance shows no advantage over directly truncating context, which does not meet our expectations.

\section{\ExtAgents: A Scalable Solution}
\label{sec:extagents}

% To systematically address the scalability bottlenecks identified in existing LLM-based multi-agent methods (Section~\ref{sec:review}), we introduce \textbf{\ExtAgents}, a novel multi-agent framework specifically designed for effective and scalable inference-time knowledge integration beyond context windows (Figure~\ref{fig:method}). The primary objectives of \ExtAgents are: 
% (i) to process massive input within limited context windows,
% (ii) to achieve performance equivalent to or surpassing that of a hypothetical infinite-context LLM,
% and (iii) to maintain efficiency through high parallelism. 

\subsection{Agent Profiles}

We adopt the distributional paradigm to partition the full input into agent-specific context chunks. 
\ExtAgents simplifies agent roles into two, corresponding to the two-stage orchestration, compatible to process any amount of input: 
\begin{itemize}[leftmargin=1.5em, nosep]
\item \textbf{Seeking Agents:} Comprehending assigned knowledge chunks and rating the relevance of a context chunk to the task query. Following Section~\ref{sec:review}, the number of Seeking Agents equals to chunk numbers, i.e., $N$.  
Optionally, they can exclude redundant chunks. 
\item \textbf{Reasoning Agent:} Integrating knowledge accumulated from Seeking Agents to generate the final answer. 
Reasoning Agent identifies the answerability and could refuse to answer if the provided information is insufficient. 
It is compatible for both multi-hop QA and long generation with switched task prompts. 
\end{itemize}

\begin{figure*}[t]
    \centering
    \includegraphics[width=\linewidth]{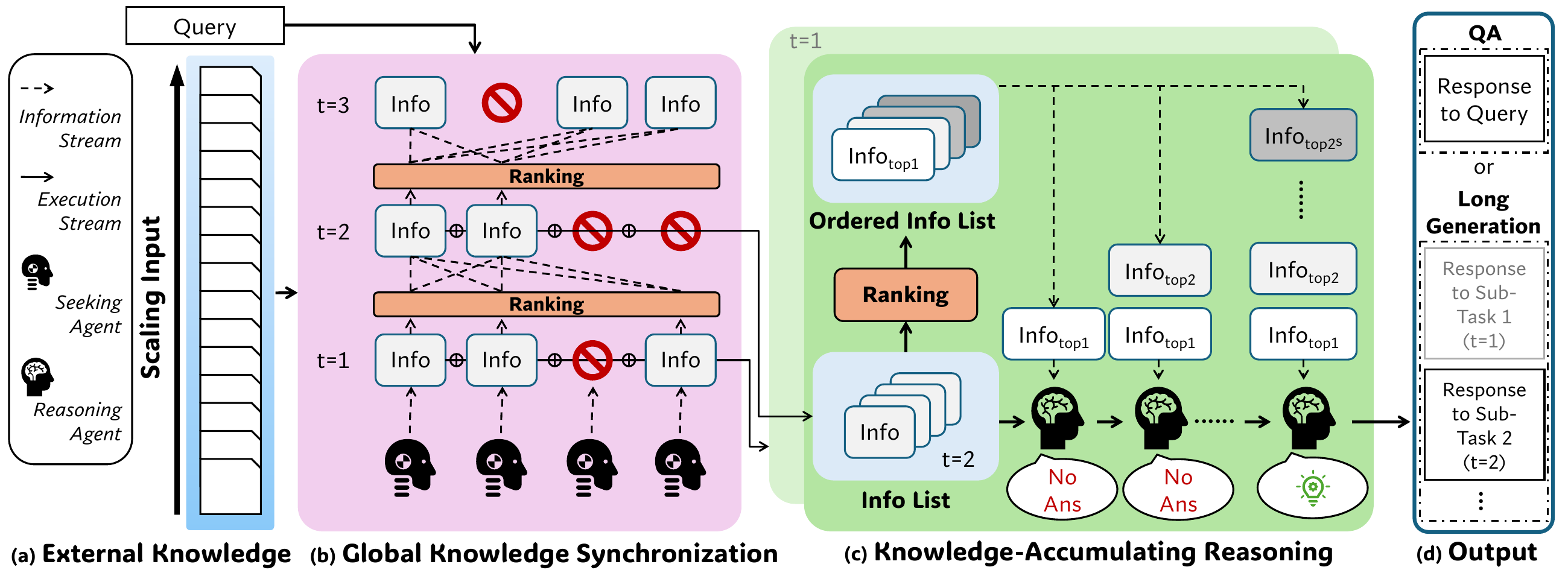}
    \caption{\textbf{Overview of \ExtAgents}: Our framework consists of multiple agents with fixed context windows, that collaboratively process (a) scalable external knowledge inputs beyond the context limit. It features (b) global knowledge synchronization, and (c) knowledge-accumulate reasoning processes by sharing a ranking mechanism at each timestep. Moreover, \ExtAgents support (d) both multi-hop QA and long survey generation tasks.}
    \label{fig:method}
    % \vspace{-1.5em}
\end{figure*}

\begin{figure*}[!t]
    \vspace{-0.5em}
    \centering
    \includegraphics[width=0.85\linewidth]{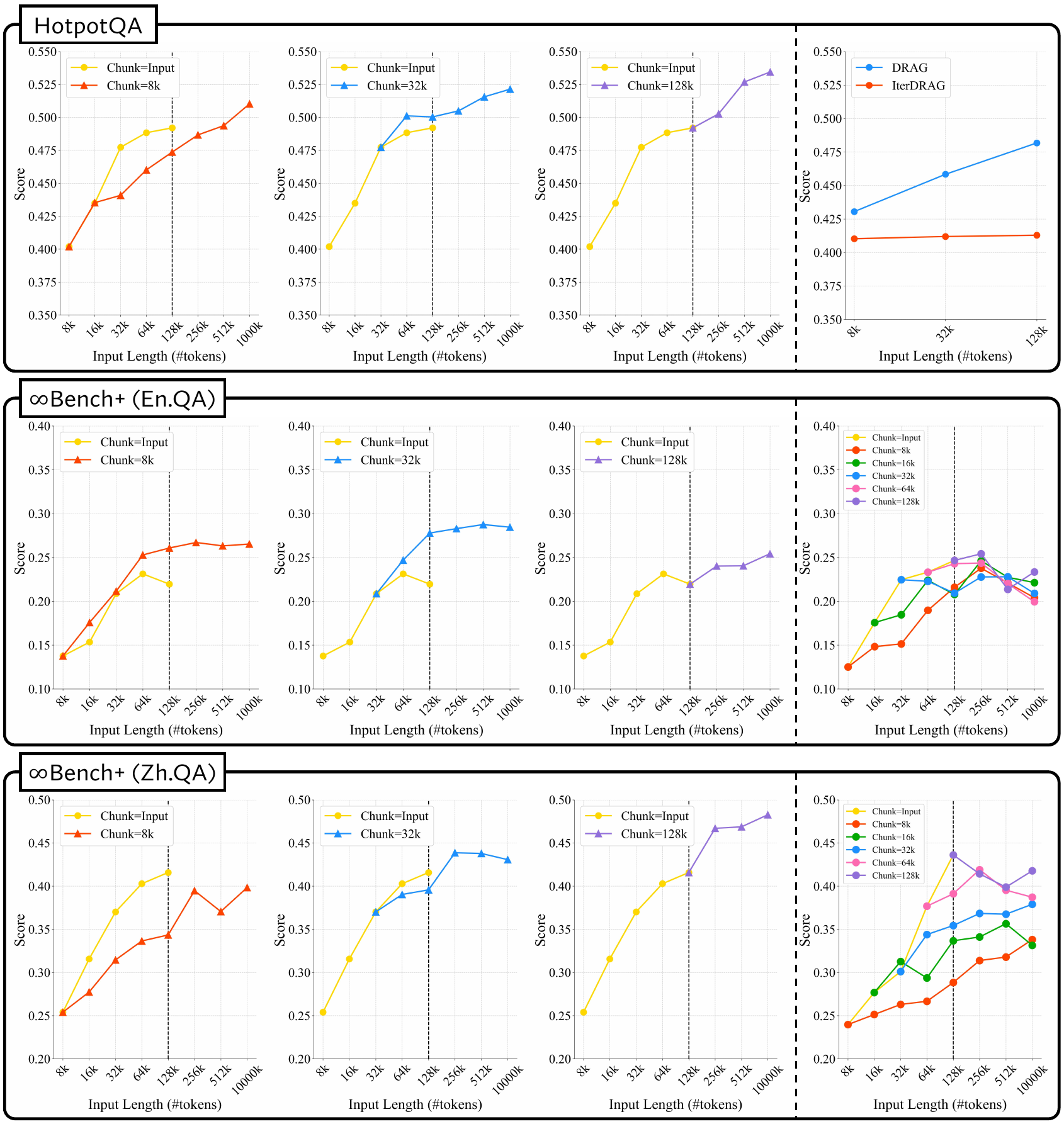}
    \vspace{-0.5em}
    \caption{Experiment of scaling external knowledge input on multi-hop QA tasks. We plot Llama-3.1-8B-Instruct results for En.QA subset in $\infty$Bench+ and gpt-4o-mini results for other tasks. The \emph{rightmost} subfigures are baseline results, including DRAG and IterDRAG on HotpotQA and LLM$\times$MapReduce on $\infty$Bench+. ``Chunk=Input'' represents that the entire context truncated is fed directly into the LLM without any chunking according to the input length. Conversely, the ``Chunk=$N$'' labels indicate that the input sequence was segmented into fixed-size chunks of length $N \in \{8\text{k}, 16\text{k}, \dots\}$ tokens.} 
    \label{fig:scaling-exp}
    \vspace{-1em}
\end{figure*}

\begin{table*}[t!]
    \centering
    \begin{minipage}{0.655\linewidth}
    \centering
    \resizebox{0.78\linewidth}{!}{
    \begin{tabular}{lcccccc}
    \toprule
    \multirow{2}{*}{\textbf{Method}} & \multicolumn{2}{c}{\textbf{HotpotQA}} & \multicolumn{2}{c}{\textbf{En.QA}} & \multicolumn{2}{c}{\textbf{Zh.QA}} \\ \cmidrule(lr){2-3} \cmidrule(lr){4-5} \cmidrule(lr){6-7}
     &  F1 &  Input & F1 &  Input &  F1 &  Input \\
    \midrule\midrule
    \multicolumn{7}{c}{\textit{DeepSeek-R1-Distill-Llama-8B}} \\
    \textcolor{gray}{Direct Input} & \textcolor{gray}{.159}  & \textcolor{gray}{32k} &  \textcolor{gray}{.097} &  \textcolor{gray}{32k} &  \textcolor{gray}{.143} &  \textcolor{gray}{32k} \\
    \midrule
    \multicolumn{7}{c}{\textit{gpt-4o-mini-2024-07-18}} \\
    Direct Input &  .204 &  128k &  .182 &  128k &  .204 &  128k \\
    DRAG &  .482 &  128k  & - & - & - & -\\
    IterDRAG &  .413 &  128k  & - & - & - & -   \\
    LLM$\times$MapReduce & - & - &  .374 &  128k &  .436 &  128k \\
    \textbf{\ExtAgents} (Ours) &  \textbf{.534} &  \textbf{1024k} &  \textbf{.382} &  \textbf{1024k} &  \textbf{.482} &  \textbf{1024k} \\
    \midrule
    \multicolumn{7}{c}{\textit{Llama-3.1-8B-Instruct}} \\
    Direct Input &  .254 &  128k &  .237 &  128k &  .315 &  128k \\
    DRAG &  .349 &  32k & - & - & - & - \\
    IterDRAG &  .368 &  32k &  - & - & - & - \\
    Chain of Agents &  -  &  -  & .168  &  32k  &  .246  &  32k  \\
    LLM$\times$MapReduce & - & -  &  .254 &  256k &  .345 &  128k \\
    \textbf{\ExtAgents} (Ours) &  \textbf{.412} &  \textbf{1024k} &  \textbf{.291} &  \textbf{1024k} &  \textbf{.347} &  \textbf{256k} \\
    \midrule
    \multicolumn{7}{c}{\textit{gpt-4o-2024-08-06}} \\
    \ExtAgents ($N=1$) &  .553 &  128k & - & - & - & - \\
    \textbf{\ExtAgents} &  \textbf{.597} &  \textbf{1024k} & - & - & - & -  \\
    \bottomrule
    \end{tabular}
    }
    \caption{Performance on Multi-Hop QA tasks with the optimal setting (chunk sizes and input lengths) and the corresponding input length (\#tokens).}
    \label{tab:qa}
    \end{minipage}
    \hfill
    \begin{minipage}{0.32\linewidth}
    \centering
    \includegraphics[width=\linewidth]{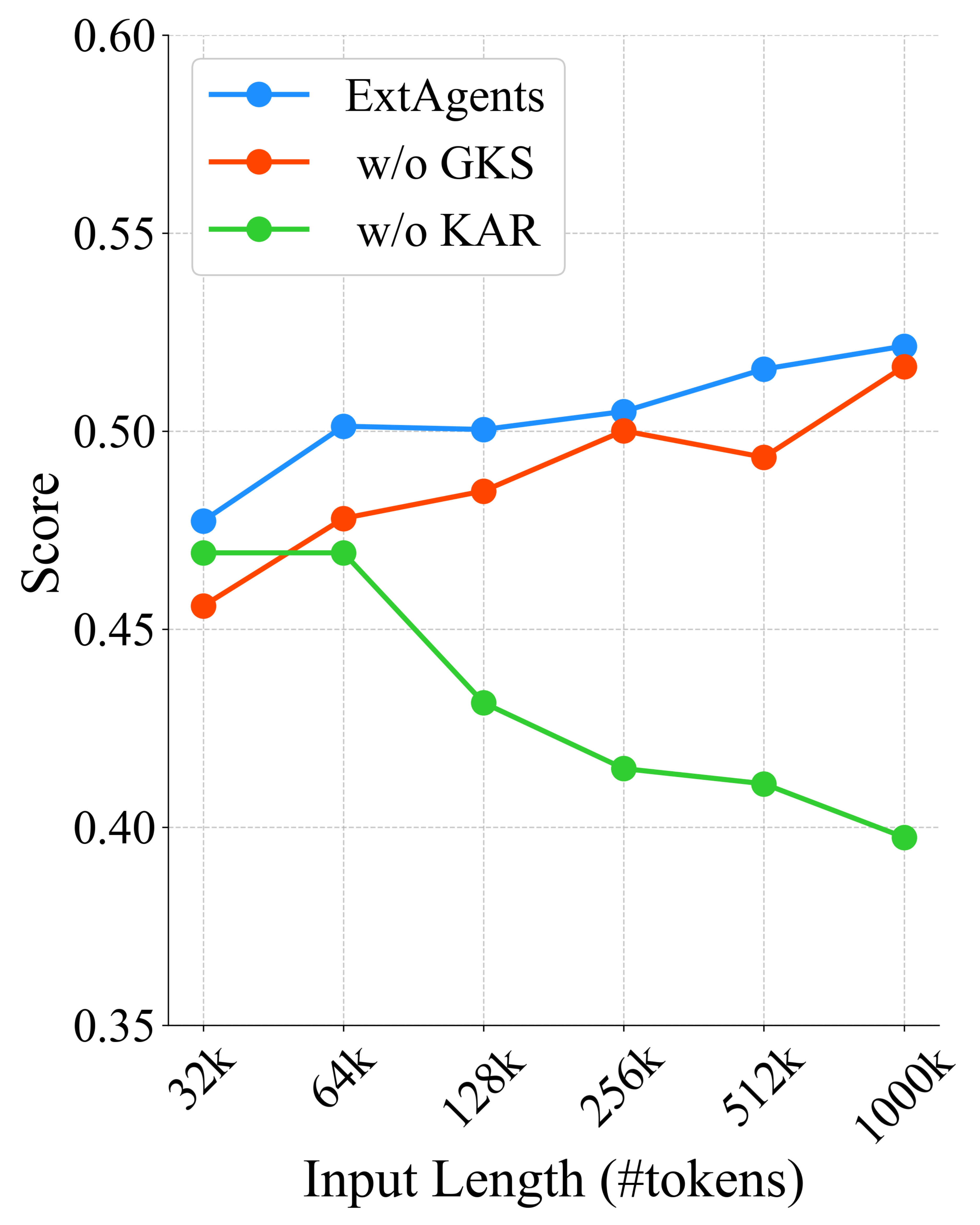}
    \captionof{figure}{Ablation studies on the global knowledge synchronization (GKS) and knowledge-accumulating reasoning (KAR) on Hotpot QA with gpt-4o-mini.}
    \label{fig:abl}
    \end{minipage}
    % \vspace{-1.5em}
\end{table*}

\subsection{Global Knowledge Synchronization}

% high parallelism alike LLM$\times$MapReduce

% Existing methods synchronize knowledge locally with limited agent interaction bandwidth, potentially restricting performance (Section~\ref{sec:review}). 
To overcome limited agent interaction bandwidth (Section~\ref{sec:review}), \ExtAgents implements \textbf{global knowledge synchronization}. %Each Seeking Agent summarizes its local context chunk into succinct messages and posts them onto a shared scratchpad accessible by all other agents. 
Unlike previous methods~\citep{zhao-etal-2024-longagent,zhou2024llmtimesmapreducesimplifiedlongsequenceprocessing}, which restrict agent interactions to local neighborhoods, our approach grants every agent global visibility by ranking messages before context assembling, thus maximizing synchronization bandwidth and ensuring propagation of salient information. 

Formally, each Seeking Agent $a_{i,t}$ at synchronization timestep $t$ updates its message as:
\vspace{-0.5em}
\begin{equation}
\scalebox{0.82}{$
m_{i,t} = a_{i,t}^{\mathrm{(EA)}}(q, d_i, \mathcal{M}_{t-1}),\quad \mathcal{M}_{t-1} = \{m_{j,t-1}\}_{j=1}^{N}, 
$}
\vspace{-0.5em}
\end{equation}
where $\mathcal{M}_{t-1}$ represents the global set of messages from the previous timestep, and $\mathrm{EA}$ stands for \ExtAgents. % This ensures comprehensive visibility and enhances collective comprehension. 
However, when the number of Seeking Agents ($N$) is large, the amount of information exchanged can break the context window. To mitigate this, each agent $a_{i,t}$ rates the relevance of the message $m_{i,t}\in\mathcal{M}_{t}$ to the task query first, outputing scores $h_{i,t} \in \mathbb{R}_{+}, i=1,...,N$. The rating could be done by appending a prompt, or by using a separate metric tool, e.g., retrieval scores. The rating method is evenly designed for each message by using the same rating principle for LLMs or the same retriever, and is performed distributedly since putting all messages together could be enormous. 
\vspace{-0.5em}
\begin{equation}
    \mathrm{Top}_k\left(\mathcal{M}_{t}\right)
    = \operatorname*{arg\,max}_{\substack{\hat{\mathcal{M}_{t}} \subseteq \mathcal{M}_{t} \\ \lvert \hat{\mathcal{M}_{t}}\rvert = k}}
    \sum_{j' \in \{j \mid m_{j,t} \in \hat{\mathcal{M}_t}\}} h_{j',t}.
\vspace{-0.5em}
\end{equation}
By selecting the top-$k$ pertinent messages with $k$ as large as possible, we maintain the global bandwidth and fit within the context window: % for knowledge synchronization at each timestep: 
\vspace{-0.5em}
\begin{multline}
m_{i,t} = a_{i,t}^{\mathrm{(EA)}}(q, d_i, \mathrm{Top}_k(\mathcal{M}_{t-1})) 
\\
\text{with}\quad 
\lvert q \rvert + \lvert d_i \rvert + \lvert \mathrm{Top}_k(\mathcal{M}_{t-1})\rvert < L. 
\vspace{-1em}
\end{multline}

\vspace{-0.5em}
Alike LLM$\times$MapReduce, all Seeking Agents can run in parallel at each timestep, substantially reducing latency with high parallelism.

\begin{table*}[t]
  \begin{minipage}[t]{0.7\linewidth}
    \vspace{0pt}
    \centering
    \resizebox{\linewidth}{!}{
    \begin{tabular}{lcccc}
      \toprule
      \textbf{Benchmark} & \textbf{LLM-as-a-Judge} ($1\sim10$) &  \textbf{\#Citations} & \textbf{Citation Density} & \textbf{Duplication Rate} \\
      \midrule\midrule
      AutoSurvey &  6.75   &   113  &  1.00   &   2.41  \\
      \textbf{\ExtAgents} (Ours) & \textbf{7.63}  &  \textbf{191}   & \textbf{1.09}  &  \textbf{1.80}  \\
      \bottomrule
    \end{tabular}
    }
    \caption{Experimental results on long survey generation tasks with gpt-4o-mini.}
    \label{tab:long-survey}
    % \resizebox{0.85\linewidth}{!}{
    %   \begin{tabular}{lcc}
    %     \toprule
    %     Method & \#Processes & Latency (s) \\
    %     \midrule\midrule
    %     Direct Input &  \phantom{0}1 &  17.80 \\
    %     \textbf{\ExtAgents} & \textbf{\phantom{0}1}  &  \textbf{19.08}  \\
    %     \multirow{2}{*}{(\textit{estimated})} &  \textbf{\phantom{0}4} &  \textbf{\phantom{0}4.94}   \\
    %     &  \textbf{16}   &   \textbf{\phantom{0}1.41}  \\
    %     \bottomrule
    %   \end{tabular}
    % }
    % \caption{Latency analysis of \ExtAgents under different numbers of processes.}
    % \label{tab:latency}
    \includegraphics[width=0.94\linewidth]{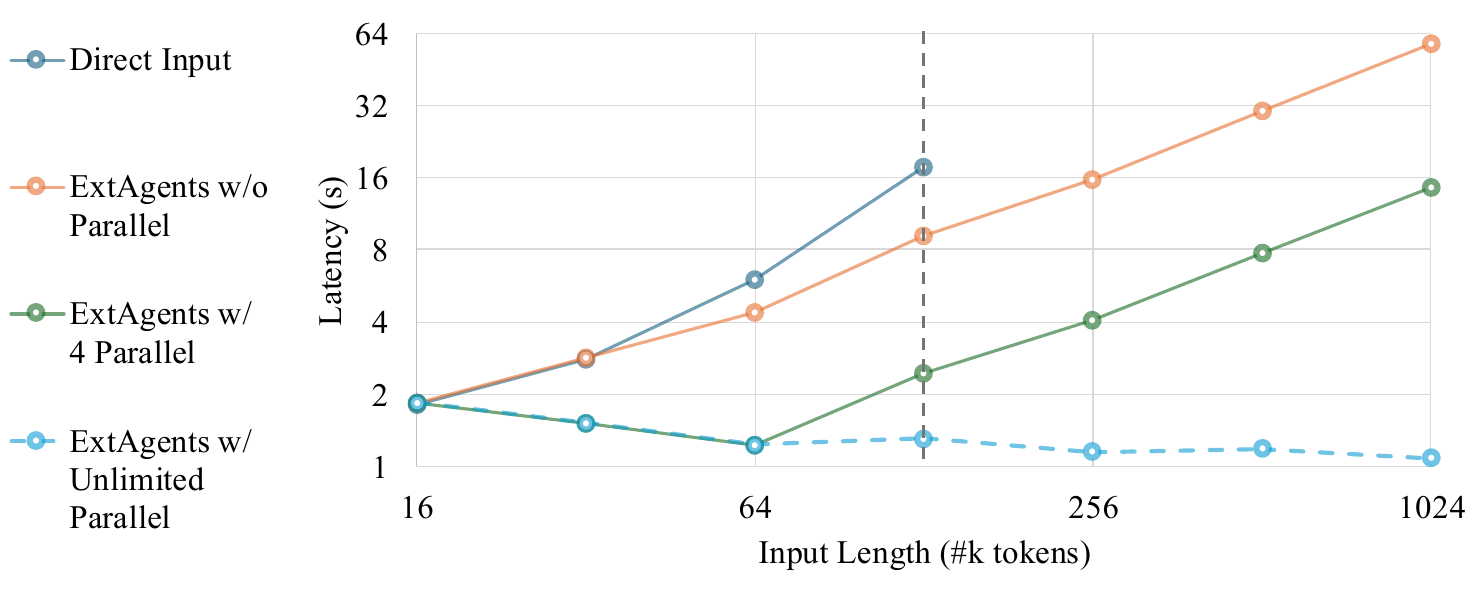}
    \vspace{-0.5em}
    \captionof{figure}{Latency analysis of \ExtAgents with the chunk size of 16k tokens.}
    \label{fig:latency}
  \end{minipage}
  \hfill
  \begin{minipage}[t]{0.28\linewidth}
    \vspace{0pt}
    \centering
    \includegraphics[width=0.95\linewidth]{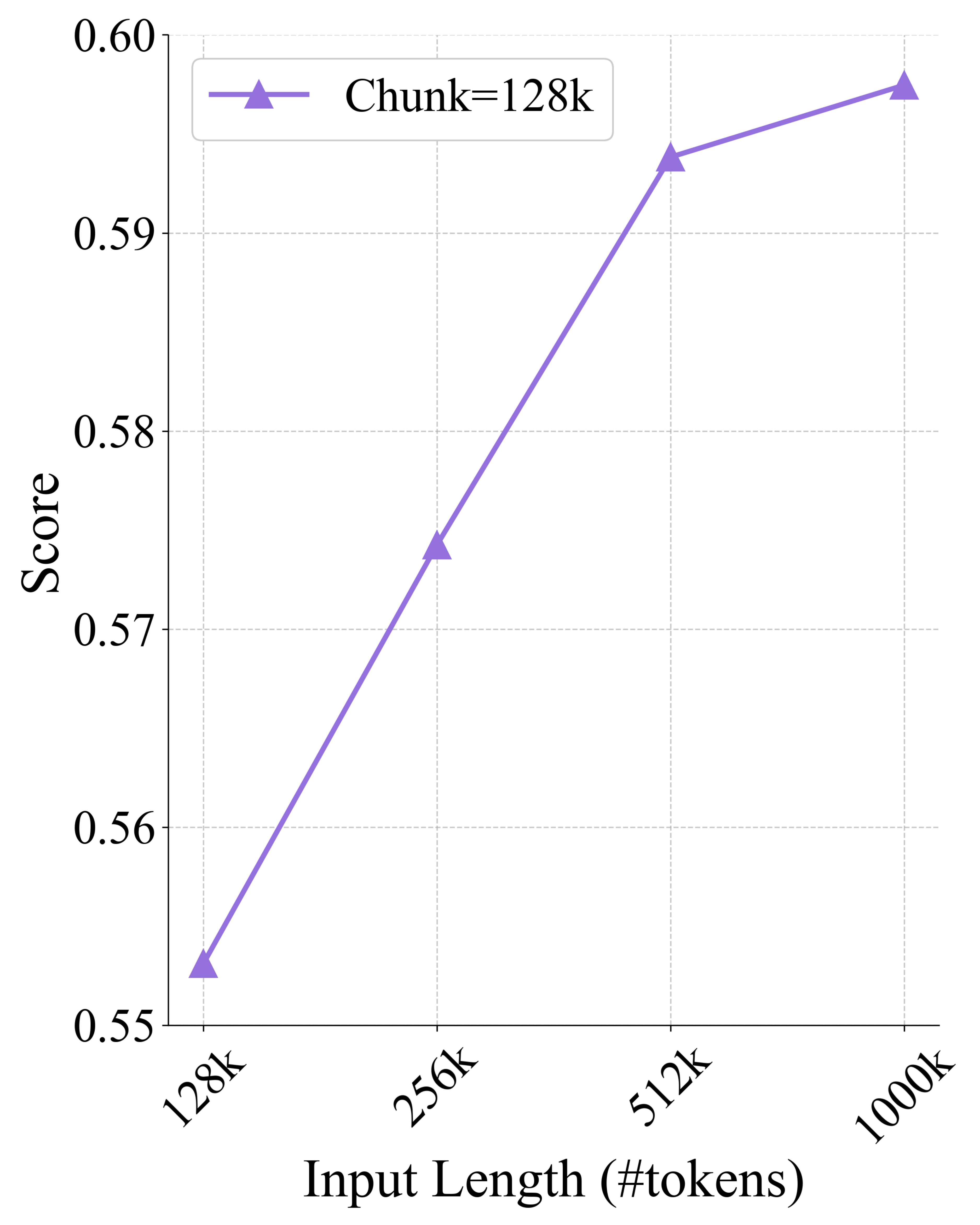}
    \captionof{figure}{Results of \ExtAgents with gpt-4o-2024-08-06 on HotpotQA benchmark.}
    \label{fig:stronger-llm}
  \end{minipage}
  % \vspace{-1.5em}
\end{table*}

\vspace{-0.5em}
\subsection{Knowledge-Accumulating Reasoning}

% TODO: #turns in average
% For long survey generation

After \emph{each knowledge synchronization timestep}, the reasoning process is initiated. 
To tackle the reasoning bottleneck caused by redundant information overload, Reasoning Agent incrementally integrates the most pertinent messages. Formally, at each reasoning iteration $s$ ($1 \le s \le S$), Reasoning Agent selects the top-$2^s$ messages from Seeking Agents at the intermediate synchronization timestep $t^\star$. The accumulated context is defined as:
\vspace{-1em}
\begin{equation}
\mathcal{M}_r^{(s)} = \mathrm{Top}_{2^s}\left(\{m_{i,t^\star}\}_{i=1}^{N}\right). 
\vspace{-0.5em}
\end{equation}
The reasoning process is performed under Equation~\eqref{eq:reason}, but Reasoning Agent first checks the answerability of the query based on given information $\mathcal{M}_r^{(s)}$ and will only output the answer if the query is answerable, which then halts the whole process of \ExtAgents.  
This iterative reasoning ensures that the Reasoning Agent progressively benefits from increased context without being overwhelmed. 
For QA tasks, the whole process is terminated when the answer is produced; or the maximum number of iterations $S$ is reached without an answer, which then starts a new round of knowledge synchronization. For long survey generation, we adopt the drafting method~\citep{wang2024autosurvey}, where the outline is generated first, and then the reasoning process is performed to fill in each section. In this case, after filling up a section, the newly started process will take the previous section into the task query for continuous generation.  

The separation of Seeking and Reasoning Agents in \ExtAgents inherently enables parallelism. When using a separate tool for rating messages, each iteration of the reasoning process could independently select synchronized messages, which can also exploit parallel computation. 
For instance, the top $(2^{s-1}+1)\sim2^s$ messages can be synchronized in parallel to the reasoning process on the top $2^{s-1}$ messages, forming an interleaved asynchronous pipeline. 
Consequently, \ExtAgents maintains high parallelism with scalability.

\section{Experiments}
\label{sec:experiments}

\vspace{-0.5em}
\subsection{Settings}
\label{sec:exp-settings}

% \looseness=-1 
\paragraph{Benchmarks} (i) \textbf{$\infty$Bench+}, our enhanced multi-hop QA benchmark with bi-lingual long documents attached to each query (Section~\ref{sec:challenge}), featuring Zh.QA and En.QA subsets, (ii) \textbf{HotpotQA}~\citep{yang-etal-2018-hotpotqa}, containing open-domain multi-hop queries related to Wikipedia, and (iii) \textbf{AutoSurvey}~\citep{wang2024autosurvey}, generating long surveys with pre-retrieved papers, as a real-world application experiment. 
Metrics include F1 for multi-hop QA and LLM-as-a-Judge for generated surveys. 
% Details are in Appendix~\ref{app:implementation}.

\paragraph{Methods}
On multi-hop QA tasks, we compare \ExtAgents with (i) \textbf{Direct Input}, the baseline method that directly inputs the truncated context into LLMs, 
(ii) \textbf{LLM$\boldsymbol{\times}$MapReduce}~\citep{zhou2024llmtimesmapreducesimplifiedlongsequenceprocessing}, the state-of-the-art multi-agent method for long-context tasks, (iii) \textbf{Chain of Agents}~\citep{zhang2024chain}, processing information in a linear topology where the bandwidth is 2, 
(iv) \textbf{DRAG} and (v) \textbf{IterDRAG}~\citep{yue2025inferencescalinglongcontextretrieval}, inference-time scalable retrieval methods for multi-hop QA with external knowledge bases. Notably, ``Direct Input'' is different from multi-agent methods without chunking, i.e., chunk size equals the input length (``Chunk$=$Input''), because of prompt templates and workflow designs. In the long survey generation application experiment, we compare \ExtAgents with \textbf{AutoSurvey}~\citep{wang2024autosurvey}, only substituting the generation process in the pipeline for fair comparison. Concurrent works~\citep{wang2025llmtimesmapreducev2entropydrivenconvolutionaltesttime, yan-etal-2025-surveyforge} propose survey generation methods with task-specific techniques, including skeleton evolving, heuristic generation, etc., which we decide is not directly comparable to our method. Details are in Appendix~\ref{app:implementation}. 

% We use the same LLM as the backbone model for all methods, including gpt-4o-mini-2024-07-18~\citep{openai2024gpt4ocard} and Llama-3.1-8B-Instruct~\citep{grattafiori2024llama3herdmodels}. 
% We test these methods with the optimal configuration except for the input length and multi-agent chunk size. For stable reproduction, we report the median results of three runs. For HotpotQA, we use BM25 retriever~\citep{10.1561/1500000019}, and for AutoSurvey, we use the original retrieval method. 
% Other details are elaborated in Appendix~\ref{app:implementation}. 

\vspace{-0.5em}
\subsection{Results and Analysis}\label{sec:results-anal}

\paragraph{Performance on Multi-Hop QA}
We plot the experimental results of scaling external knowledge input on multi-hop QA tasks in Figure~\ref{fig:scaling-exp}. Detailed illustration is in Appendix~\ref{app:implementation}. The increasing trend of performance w.r.t.~the input length indicates the scalability of each context window extension method. Empirically, \ExtAgents consistently outperforms the baselines across all input lengths, achieving the significantly better performance on both HotpotQA and $\infty$Bench+ benchmarks. However, optimal chunk sizes are subjective to different tasks and require tuning. Moreover, the performance consistently improves with the increase of external knowledge input, demonstrating the scalability of \ExtAgents. 
We also summarize the performance and input length of each method within its optimal setting in Table~\ref{tab:qa}. \ExtAgents achieves the best performance on all three multi-hop QA benchmarks by effectively utilizing more external knowledge compared to other methods. 
As another paradigm of inference-time scaling, long reasoning chains do not benefit as much (see DeepSeek-R1-Distill-Llama-8B line).

% \begin{table*}
% \centering
%   \resizebox{0.7\linewidth}{!}{
%   \begin{tabular}{lcccc}
%     \toprule
%     Benchmark & LLM-as-a-Judge ($1\sim10$) &  \#Citations & Citation Density & Duplication Rate \\
%     \midrule\midrule
%     AutoSurvey &  6.75   &   113  &  1.00   &   2.41  \\
%     \textbf{\ExtAgents} (Ours) & \textbf{7.63}  &  \textbf{191}   & \textbf{1.09}  &  \textbf{1.80}  \\
%     \bottomrule
%   \end{tabular}
%   }
%   \caption{Experimental results on long survey generation tasks.}
%   \label{tab:long-survey}
%   \vspace{-1.5em}
% \end{table*}

\begin{table*}[t]
\centering
\resizebox{0.77\linewidth}{!}{
\begin{tabular}{lcc}
\toprule
\textbf{Model Configuration} & \textbf{F1} & \textbf{Average Time per Sample} \\
\midrule
Llama-3.1-8B (Both Roles) & .383 & 115.87 \\
Llama-3.2-3B (Both Roles) & .328 & \phantom{0}36.06 \\ \midrule
Llama-3.2-3B (Seeking) + Llama-3.1-8B (Reasoning) & .363 & \phantom{0}36.52 \\
\bottomrule
\end{tabular}
}
\caption{Comparison of performance and inference time (s) across homogeneous and heterogeneous model configurations in different agent roles of \ExtAgents on HotpotQA. Input length and chunk size are fixed to 512k and 32k, respectively. Time is measured without parallelization on 4 A100 GPUs.}
\label{tab:model-mix}
\end{table*}

\paragraph{Performance on Long Survey Generation} % llama 3.1 8B
% \begin{wraptable}[7]{r}{0.5\linewidth}
%   \centering
%   \vspace{-1em}
%   \caption{Experimental results on long survey generation tasks.}
%   \label{tab:long-survey}
%   \resizebox{\linewidth}{!}{
%   \begin{tabular}{lcccc}
%     \toprule
%     Benchmark & LLM-as-a-Judge ($1\sim10$) &  \#Citations & Citation Density & Duplication Rate \\
%     \midrule\midrule
%     AutoSurvey &     &     &     &     \\
%     \textbf{\ExtAgents} & \textbf{}  &  \textbf{}   & \textbf{}  &  \textbf{}  \\
%     \bottomrule
%   \end{tabular}
%   }
% \end{wraptable}
We test \ExtAgents on long survey generation as a real-world application. \ExtAgents could incorporate more related papers, and achieves better performance with more citations and lower duplication rate compared to AutoSurvey (Table~\ref{tab:long-survey}). 
By aggregating eight pairwise scores from LLM-as-a-Judge, we find that \ExtAgents achieves a higher quality score with a significant margin. However, the evaluation of long surveys is challenging even for human experts. Thus, we also include a part of generated texts in Appendix~\ref{app:qualitative} for qualitative comparisons besides designed metrics. 
Overall, it indicates that \ExtAgents helps generate long surveys with higher quality and lower redundancy. 

% \begin{wraptable}[9]{r}{0.42\linewidth}
%   \centering
%   \vspace{-1em}
%   \caption{Latency analysis of \ExtAgents under different numbers of processes.}
%   \label{tab:latency}
%   \resizebox{\linewidth}{!}{
%   \begin{tabular}{lcc}
%     \toprule
%     Method & \#Processes & Latency (s) \\
%     \midrule\midrule
%     Direct Input &  \phantom{0}1 &  17.80 \\
%     \textbf{\ExtAgents} & \textbf{\phantom{0}1}  &  \textbf{19.08}  \\
%     \multirow{2}{*}{(\textit{estimated})} &  \textbf{\phantom{0}4} &  \textbf{\phantom{0}4.94}   \\
%      &  \textbf{16}   &   \textbf{\phantom{0}1.41}  \\
%     \bottomrule
%   \end{tabular}
%   }
% \end{wraptable}

% \begin{table}
%   \centering
%   \resizebox{\linewidth}{!}{
%   \begin{tabular}{lcc}
%     \toprule
%     Method & \#Processes & Latency (s) \\
%     \midrule\midrule
%     Direct Input &  \phantom{0}1 &  17.80 \\
%     \textbf{\ExtAgents} & \textbf{\phantom{0}1}  &  \textbf{19.08}  \\
%     \multirow{2}{*}{(\textit{estimated})} &  \textbf{\phantom{0}4} &  \textbf{\phantom{0}4.94}   \\
%      &  \textbf{16}   &   \textbf{\phantom{0}1.41}  \\
%     \bottomrule
%   \end{tabular}
%   }
%   \caption{Latency analysis of \ExtAgents under different numbers of processes.}
%   \label{tab:latency}
% \end{table}

\paragraph{Latency and Cost Analysis}
In HotpotQA benchmark, we measure the latency of \ExtAgents with fixed 16k-token chunks under different amounts of input with Llama-3.1-8B-Instruct on 4 A100 GPUs (Figure~\ref{fig:latency}). 
The latency of direct input grows quadratically with input length, while \ExtAgents maintains a linear growth for a fixed number of parallel threads. The theoretical complexity analyses are similar to Chain of Agents~\citep{zhang2024chain}. 
Under unlimited parallel threads, ideally, \ExtAgents could even decrease the latency because of fully paralleled Seeking Agents and the interleaved asynchronous pipeline (Section~\ref{sec:extagents}).

\paragraph{Ablation Studies}
We conduct ablation studies on the global knowledge synchronization (GKS) and knowledge-accumulating reasoning (KAR) in \ExtAgents to analyze the identified bottlenecks (Figure~\ref{fig:abl}). 
The results show that removing KAR leads to a significant drop in performance, especially as the amount of external knowledge increases. This demonstrates that the KAR effectively breaks the bottleneck of information overload. 
Removing GKS also leads to a drop in different input lengths, demonstrating that leveraging the range of exchanged information (i.e.~``bandwidth'') could help catch more relevant information.

% \begin{wrapfigure}[17]{r}{0.6\linewidth}
%     \centering
%     \includegraphics[width=\linewidth]{./figs/Figures_PDF/RAG_gpt-4o/f1_4o_v3_128k.pdf}
%     \caption{Results of \ExtAgents with gpt-4o-2024-08-06 on HotpotQA benchmark with strong scalability.}
%     \label{fig:stronger-llm}
% \end{wrapfigure}

% \begin{figure}
%     \centering
%     \includegraphics[width=0.6\linewidth]{./figs/Figures_PDF/RAG_gpt-4o/f1_4o_v3_128k.pdf}
%     \caption{Results of \ExtAgents with gpt-4o-2024-08-06 on HotpotQA benchmark with strong scalability.}
%     \label{fig:stronger-llm}
%     \vspace{-1.5em}
% \end{figure}

\paragraph{Compatibility across LLM Families}
% Better model better
We further test \ExtAgents with gpt-4o, and find that the performance is more significantly improved with the stronger LLM (Figure~\ref{fig:stronger-llm}) compared to weaker models (Figure~\ref{fig:scaling-exp}). Moreover, Llama-3.1-8B-Instruct suffers linguistic bias to achieve better performance on En.QA than Zh.QA, while gpt-4o-mini performs consistently well on both subsets (Table~\ref{tab:qa}), affecting Llama's scalability on QA tasks in Chinese and with code-mixed knowledge bases (Figure~\ref{fig:scaling-exp-llama}). 
Further analyses are in Appendix~\ref{app:implementation}. 
% The reason might be that the stronger LLMs are more capable of collaboration and comprehension. 
Overall, \textbf{stronger LLMs benefit more from the scalability of \ExtAgents}, implying promising future work with even stronger LLMs.

\paragraph{Heterogeneous Model Collaboration for Enhanced Efficiency}
A primary advantage of the decoupled agent profiles in ExtAgents is the flexibility to assign heterogeneous models to different roles based on their computational requirements. To explore the efficiency and performance trade-offs, we investigate whether a smaller, highly efficient model can serve as the Seeking Agent while a stronger model handles the more complex reasoning tasks. We conducted experiments pairing Llama-3.2-3B-Instruct for Seeking Agents with Llama-3.1-8B-Instruct for the Reasoning Agent on HotpotQA benchmark. Inference time was measured without parallelism on 4 A100 GPUs. For the heterogeneous setting, the two LLMs are deployed on 2 GPUs, respectively. As shown in Table \ref{tab:model-mix}, offloading the extensive synchronization and extraction workloads to the smaller model improves efficiency, achieving a \textasciitilde $3.1\times$ speedup compared to using the 8B model for both roles, matching the computational efficiency of the pure 3B setup while delivering superior task performance.

% \vspace{-0.5em}
\section{Conclusion and Future Work}
\label{sec:conclusion}

We introduce \textbf{\ExtAgents}, a multi-agent framework that scales external knowledge input beyond context windows of LLMs  without additional training. By decoupling knowledge synchronization and reasoning, \ExtAgents overcomes core bottlenecks of existing methods, consistently improving multi-hop QA and long-form generation performance while maintaining high parallel efficiency. Future work will explore adaptive agent orchestration, cross-modal and tool augmentation, and fine-tuning specialized models.% toward more scalable knowledge-centric agents.

\section*{Limitations}

\label{app:limitations}

\paragraph{Model Alignment}
\ExtAgents inherits both the strengths and weaknesses of its underlying LLMs: while substantial scalability on external knowledge input has shown in \ExtAgents, the framework offers no principled defense on adversarial models, e.g., the aggregated evidence might be factually incorrect, biased, or policy-incompliant. Misaligned or adversarial Seeking Agents can propagate errors to every Reasoning Agent, amplifying harmful content or systemic biases. Incorporating alignment-aware scoring, preference-based post-training, or tool-based content filters could alleviate the problem but introduce training costs, which is left to future work. 

\paragraph{Integration with Chunking Techniques}
\ExtAgents partitions long inputs into fixed-size slices to simplify agent contexts, but makes no attempt to optimize those boundaries. Advanced chunking strategies---semantic segmentation, overlap windows, or hierarchical compression---could further reduce information loss~\citep{chen-etal-2024-dense,duarte-etal-2024-lumberchunker,zhao2024metachunkinglearningefficienttext}, yet also introduce new coordination overhead and hyper-parameter choices. A systematic study of how adaptive chunking interacts with agent synchronization and reasoning quality is beyond the scope of this work. 

\paragraph{Perfect Rankings with Further Calibration} 
Though with current ranking design, the performance advancement has been empirically shown, the rankings could be unreliable in some extreme situations, e.g., out-of-domain or adversarial inputs to the rating method, rating biases in weaker LLMs or retrievers, etc. And, we indeed observe stronger LLMs could achieve better end-to-end performance in Section~\ref{sec:results-anal}, potentially partially due to its stronger judging capability. We suggest further calibration methods to enhance the rating reliability in future work: (1) Calibration with labelled validation sets for fitting retriever scores, in-context learning in LLMs, etc.; (2) Calibration with multiple votes~\citep{lin2024generating}; (3) Rating with stronger or specifically trained models.

\section*{Ethical Considerations}

\ExtAgents should be viewed as a step towards more scalable and efficient knowledge-centric LLM workflow. Thus, different domains may be impacted either positively or negatively depending on the specific use case.  
For instance, in the educational domain, our work could enhance personalized learning and research productivity by allowing teachers and students to access and reason over extensive knowledge beyond textbooks, potentially democratizing expert-level insights and reducing barriers to advanced inquiry. However, there might also be negative implications: the improved scalability in integrating large-scale external knowledge may unintentionally amplify misinformation or biased viewpoints, as automated retrieval and reasoning processes could propagate inaccuracies present in the underlying data sources, especially for medical or economical industries. To mitigate these risks, further development of verification mechanisms and post-training techniques to align agent-produced knowledge is recommended.

\section*{Acknowledgments}

This work is supported by the National Natural Science Foundation of China (No. 62276152, 62236011), Key Laboratory of Ethnic Language Intelligent Analysis and Security Governance of MOE, Minzu University of China, Beijing, China. We thank Shuo Wang and Shengding Hu for the discussion during the early development of this project.

% \begin{figure}
%   \centering
%   \fbox{\rule[-.5cm]{0cm}{4cm} \rule[-.5cm]{4cm}{0cm}}
%   \caption{Sample figure caption.}
% \end{figure}

% \begin{table}
%   \caption{Sample table title}
%   \label{sample-table}
%   \centering
%   \begin{tabular}{lll}
%     \toprule
%     \multicolumn{2}{c}{Part}                   \\
%     \cmidrule(r){1-2}
%     Name     & Description     & Size ($\mu$m) \\
%     \midrule
%     Dendrite & Input terminal  & $\sim$100     \\
%     Axon     & Output terminal & $\sim$10      \\
%     Soma     & Cell body       & up to $10^6$  \\
%     \bottomrule
%   \end{tabular}
% \end{table}

% \begin{ack}
% Use unnumbered first level headings for the acknowledgments. All acknowledgments

% Do {\bf not} include this section in the anonymized submission, only in the final paper. You can use the \texttt{ack} environment provided in the style file to automatically hide this section in the anonymized submission.
% \end{ack}

% \bibliographystyle{acl_natbib}
\bibliography{custom}

%%%%%%%%%%%%%%%%%%%%%%%%%%%%%%%%%%%%%%%%%%%%%%%%%%%%%%%%%%%%

\clearpage

\appendix

% \section{Broader Impact}
% \label{app:impact}

% TODO: GPT-4o validation

\section{Details of Review on Existing Multi-Agent Methods}
\label{app:review-detail}

In this section, we provide the implementation of existing LLM-based multi-agent systems for context window extension, including Chain of Agents~\citep{zhang2024chain}, LongAgent~\citep{zhao-etal-2024-longagent}, and LLM$\boldsymbol{\times}$MapReduce~\citep{zhou2024llmtimesmapreducesimplifiedlongsequenceprocessing} in our framework in Section~\ref{sec:review}. We also explain the comparison results in Table~\ref{tab:existing-methods}. 

\paragraph{Knowledge Synchronization}
Here, we explain how existing methods synchronize knowledge across agents according to Equation~\eqref{eq:sync}. 
For \textbf{Chain of Agents}, each agent $a_i,t$ incorporate the message from previous agent $a_{i-1,t-1}$ in a linear topology, and the message is passed to the next agent $a_{i+1,t-1}$ in the next timestep. So in Equation~\eqref{eq:sync}, $\mathcal{D}_{\mathcal{G}_{i,t-1},t}^{\mathrm{(CoA)}} = \{d_i\}$ and $\mathcal{M}_{\mathcal{G}_{i,t-1},t-1}^{\mathrm{(CoA)}} = \{m_{i-1,t-1}\}$. Since $\mathcal{G}_{i,t-1}^{\mathrm{(CoA)}} = \{a_{i-1,t-1}, a_{i,t-1}\}$, the bandwidth is $2$. It is also clear that the process could not be parallelized. 
For \textbf{LongAgent}, the leader agent identify conflicts between two agents $a_{i,t-1}$ and $a_{j,t-1}, (j\neq i)$, and the individual message and original chunk are passed to each other. So in Equation~\eqref{eq:sync}, $\mathcal{D}_{\mathcal{G}_{i,t-1},t}^{\mathrm{(LA)}} = \{d_i, d_j\}$ and $\mathcal{M}_{\mathcal{G}_{i,t-1},t-1}^{\mathrm{(LA)}} = \{m_{i,t-1}, m_{j,t-1}\}$. Thus, $\mathcal{G}_{i,t-1}^{\mathrm{(LA)}} = \{a_{i,t-1}, a_{j,t-1}\}$, and the bandwidth is $2$. 
For each timestep, the two agents could function in parallel. 
Though, the total timesteps $T$ is $O(\frac{L}{\lvert m\rvert})$, which is the number of synchronization decisions made by the leader agent. However, due to the capacity of the leader agent, $T$ is rather small and most agents barely access all other contexts. Specifically, the task query $q$ might be changed to generated sub-queries by the leader agent during the process. 
For \textbf{LLM$\boldsymbol{\times}$MapReduce}, agents function in parallel and aggregate their messages in groups with adjacent agents. The group size is $O(\frac{L}{\lvert m\rvert})$ for agent to process messages within the context window at the next timestep. And $\mathcal{M}_{\mathcal{G}_{i,t-1},t-1}^{\mathrm{(MR)}} = \phi$. Thus, the bandwidth is $O(\frac{L}{\lvert m\rvert})$. Each group could be processed in parallel at the same timestep. 

\paragraph{Knowledge-Integrated Reasoning}
Here, we explain how existing methods integrate knowledge in reasoning according to Equation~\eqref{eq:reason}.
For \textbf{Chain of Agents}, the reasoning agent only takes the message from the last agent $a_{N,T}$ as the reasoning context $\mathcal{M}_r^{\mathrm{(CoA)}} = \{m_{N,T}\}, (T=N)$. 
For \textbf{LongAgent}, the leader agent takes all messages from all agents in sequential as the reasoning context $\mathcal{M}_r^{\mathrm{(LA)}} = \{m_{i,t}\}_{1\le i\le N, 1\le t\le T}$. Since $T = O(\frac{L}{\lvert m\rvert})$, the context would not exceed the context window. 
For \textbf{LLM$\boldsymbol{\times}$MapReduce}, the reduce agent takes the last messages from all agents as the reasoning context $\mathcal{M}_r^{\mathrm{(MR)}} = \{m_{i,T}\}_{1\le i\le N}$. The method guarantees that the reasoning context is within the context window by recursively aggregating the messages in groups with adjacent agents. 
The last two methods tend to include as much information as possible in the reasoning context, which may lead to overload with redundant, noisy information.

\section{Qualitative Analysis}
\label{app:qualitative}

For multi-hop QA tasks, we show three cases as follows. The first case demonstrates \ExtAgents' superior ability to connect disparate pieces of information across multiple documents to answer a complex, multi-step question accurately, where other methods fail to synthesize the necessary facts or get sidetracked by redundant details. The second case highlights \ExtAgents' effectiveness in pinpointing the correct answer by intelligently scoring and prioritizing the most relevant textual evidence, especially when multiple, potentially conflicting pieces of information are present within the source material. The third case shows \ExtAgents' strength in accurately identifying the main subject by discerning the most pertinent information through its scoring mechanism, successfully avoiding misdirection from other related but less central entities mentioned in the text in different languages. 

For long survey generation, we present a comparison of one subsection generated by our method and AutoSurvey for the same topic. The survey generated by \ExtAgents presents a more related and more diverse range of innovative LLM applications in education, making the overall argument more persuasive and resulting in overall 68 pages compared to 58 pages from AutoSurvey.  

\onecolumn
\subsection{Case: Multi-Hop Question Answering}

% \subsection{Example for RAG}
% The case below demonstrates \ExtAgents' superior ability to connect disparate pieces of information across multiple documents to answer a complex, multi-step question accurately, where other methods fail to synthesize the necessary facts or get sidetracked by irrelevant details.

\begin{tcolorbox}[enhanced,
  breakable, colback=cyan!10!white, colframe=cyan!40!black,, title=Example from HotpotQA: Question Answering on William Hodges' Voyage]
\textbf{Question:} William Hodges visited a polar region as part of a voyage, the region is how big?\\[5pt]
\textbf{Answer:} 14 million km\\[5pt]
\textbf{Excerpts from retrieved documents:}\\[5pt]
% \begin{quote}
William Hodges RA (28 October 1744 – 6 March 1797) was an English painter. He was a member of James Cook's second voyage to the Pacific Ocean, and is best known for the sketches and paintings of locations he visited on that voyage, including Table Bay, Tahiti, Easter Island, New Zealand, Dusky Sound and the Antarctic.\\[5pt]
% \end{quote}
% \begin{quote}
The Antarctic (US English, UK English or and or) is a polar region around the Earth's South Pole, opposite the Arctic region around the North Pole. The Antarctic comprises the continent of Antarctica, the Kerguelen Plateau and other island territories located on the Antarctic Plate or south of the Antarctic Convergence. The Antarctic region includes the ice shelves, waters, and all the island territories in the Southern Ocean situated south of the Antarctic Convergence, a zone approximately wide varying in latitude seasonally. The region covers some 20 percent of the Southern Hemisphere, of which 5.5 percent (14 million km) is the surface area of the Antarctic continent itself.\\[5pt]\\[5pt]
% \end{quote}
\textbf{\ExtAgents}\\[5pt]
Extracted information (partial):\\[5pt]
William Hodges, an English painter, accompanied James Cook on his second voyage to the Pacific Ocean between 1772 and 1775. During this voyage, he traveled to various locations, including Antarctic regions. However, the chunk does not provide the exact size of the polar region Hodges visited. It mentions that Hodges is known for his sketches and paintings of locations he visited on that voyage, including those in the Antarctic, but does not specify the size of the region.\\[5pt]
William Hodges was likely referring to the polar regions explored during his voyages. Specifically, he could relate to the Antarctic region, which is noted as a large landmass and describes being covered by about 98\% ice, averaging in thickness. The Antarctic continent itself encompasses an area of approximately 14 million square kilometers (or around 5.5\% of the Southern Hemisphere). \\[5pt]
\textbf{Prediction:} Approximately 14 million square kilometers. \\[5pt]\\[5pt]
\textbf{DRAG prediction:} The region is thought to have covered about 1.5 million square kilometers.\\[5pt]\\[5pt]
\textbf{IterDRAG prediction:} The specific size of the broader region William Hodges visited is not mentioned in the context; however, one specific area he explored, Hodges Knoll, is 2.2 km in northeast-southwest direction and 1 km wide.
% \textbf{Analysis:}
% Our method effectively handled the larger input size and combined relevant knowledge across multiple chunks. DRAG, however, made an error in size estimation due to limitations in processing large contexts. IterDRAG failed to provide the correct answer, as it could only extract limited specific information.
\end{tcolorbox}

\clearpage
% \subsection{Example for En.QA}
% The case below highlights \ExtAgents' effectiveness in pinpointing the correct answer by intelligently scoring and prioritizing the most relevant textual evidence, especially when multiple, potentially conflicting pieces of information are present within the source material. 
% \begin{figure}[H]
\begin{tcolorbox}[enhanced,
  breakable,colback=cyan!10!white, colframe=cyan!40!black, title=Example from En.QA in $\infty$Bench+: Question Answering on Davion's Fortune]
\textbf{Question:} Where is the Davion’s fortune hidden?\\[5pt]
\textbf{Answer:} On the island of Brayan Annabel\\[5pt]
\textbf{Excerpts from original text:}\\[5pt]
Davion followed him with an excited look. "and now," he said, when he saw that Pascal had read the last line, "put the two fragments together, and judge for yourself." Pascal obeyed, and the conjointed pieces gave the following: -- "This 25th day of April, 1498, be...ing invited to dine by his Holiness Alexander VI., and fearing that not...content with making me pay for my hat, he may desire to become my heir, and re...serves for me the fate of Cardinals Caprara and Bentivoglio, who were poisoned...I declare to my nephew, Guido Spada, my sole heir, that I have bu...ried in a place he knows and has visited with me, that is, in...the caves of the small Island of Brayan Annabel all I poss...ssed of ingots, gold, money, jewels, diamonds, gems; that I alone...know of the existence of this treasure, which may amount to nearly two mil...lions of Roman crowns, and which he will find on raising the twentieth ro...ck from the small creek to the east in a right line. Two open...ings have been made in these caves; the treasure is in the furthest a...ngle in the second; which treasure I bequeath and leave en...tire to him as my sole heir."\\[5pt]\\[5pt]
\textbf{\ExtAgents}\\[5pt]
Extracted information and corresponding scores (partial): \\[5pt]
Davion's fortune is hidden in a coffer located in a cave on the Island of Brayan Annabel. The coffer is described as being bound with iron and engraved with the arms of the Spada family. It contains piles of golden coins, bars of unpolished gold, and a significant amount of diamonds, pearls, and other gems. The treasure is concealed beneath a large circular rock that Pascal discovers and manages to move, revealing the coffer underneath. Score: 100.0\\[5pt]
% The Davion's fortune is hidden in a grotto at Brayan Annabel, which Brisa knows the secret of. In the context of the story, Brayan Annabel mentions that he has buried a treasure there, which he had originally intended for his betrothed before he was imprisoned. The fortune consists of a significant amount of money, specifically twenty million, which he bequeaths to Lorna Julianne, the son of his former patron, Pierre Julianne. Score: 90.0\\[5pt]
\textbf{Prediction:} Island of Brayan Annabel\\[5pt]\\[5pt]
\textbf{LLM$\times$MapReduce}\\[5pt]
Extracted information (partial): \\[5pt]
The Davion's fortune is hidden in a treasure chest buried in the caves on the Island of Brayan Annabel.\\[5pt]
The Davion's fortune is hidden in a diamond that belonged to a character named Armando Pascal, who bequeathed it to Corbin, the innkeeper, before his death.\\[5pt]
Davion's fortune is hidden in a grotto at Brayan Annabel, of which Brisa knows the secret.\\[5pt]
The Davion's fortune is hidden in a chest concealed under the closet in the little house in the Allees de Meillan, at Marseilles.\\[5pt]
\textbf{Prediction:} The Davion's fortune is hidden in multiple locations: a treasure chest in the caves on the Island of Brayan Annabel, a diamond bequeathed to Corbin, a grotto at Brayan Annabel, and a chest under a closet in a house in Allees de Meillan, Marseilles.
% \\[5pt]
% \textbf{Analysis:} The original text repeatedly mentions the location of Davion's fortune, and the Map Reduce method merges all information, leading to an imprecise answer. Our method, by scoring the information and prioritizing the most relevant chunks, produces an accurate result.
\end{tcolorbox}
% \caption{Example for En.QA}
% \end{figure}

% \subsection{Example for Zh.QA}
% The case below shows \ExtAgents' strength in accurately identifying the main subject by discerning the most pertinent information through its scoring mechanism, successfully avoiding misdirection from other related but less central entities mentioned in the text in different languages. 
% \begin{figure}[H]
\begin{tcolorbox}[enhanced,
  breakable,colback=cyan!10!white, colframe=cyan!40!black, title=Example from Zh.QA in $\infty$Bench+: Question Answering on Who Infiltrated the Scarlet Research Institute]
\textbf{Question:} 谁混入猩红研究院开启他的卧底计划\\[5pt]
\textbf{Answer:} 贾易\\[5pt]
\textbf{Excerpts from original text:}\\[5pt]
“我决定让我的学生贾易开始接触生物科学领域，希望他不会让我失望，以他的学习能力，如果努力的话，不需要太久就能精通这方面的内容，并混入猩红研究院。”\\
“贾易已经通过了猩红研究院的两轮考核，即将混入猩红研究院开启他的卧底计划，希望这次能从中得到有用的信息。”\\
“贾易已经混入猩红研究院了，我等待着他传来的好消息。”\\[5pt]\\[5pt]
\textbf{\ExtAgents}\\[5pt]
Extracted information and corresponding scores (partial): \\[5pt]
在故事中，混入猩红研究院开启卧底计划的角色是贾易。他是封棋的学生，负责保护可能被暗杀的人类天才，并试图从猩红研究院获取有用的信息。贾易的任务是潜入猩红研究院，调查其内部情况，尤其是与符文改造人计划相关的秘密。Score: 100.0\\[5pt]
在文章中，混入猩红研究院并开启卧底计划的是“黑影”。黑影是灵能族的叛徒，曾经在灵能族中拥有强大的实力。文章提到黑影在科技研究院布局未来，并且在战争线中与封棋等人产生了交集。封棋在与黑影的对抗中，逐渐意识到黑影的真实身份和目的。Score: 90.0\\[5pt]
% 根据您提供的文本，混入猩红研究院开启卧底计划的是“不明身份的势力”，可以推测这个势力与“黑影”及“科技研究院”有关。具体细节并没有直接在文本中明示，但提到星城内存在的潜伏势力，以及黑影对其的掌控，暗示了其在行动上的秘密与阴谋。因此，确切的身份并未明确，但从上下文可以推测出可能与更多复杂的领域生物或势力有关。Score: 70.0\\[5pt]
\textbf{Prediction:} 贾易\\[5pt]\\[5pt]
\textbf{LLM$\times$MapReduce}\\[5pt]
Extracted information (partial): \\[5pt]
在小说中，混入猩红研究院开启卧底计划的是“贾易”。他是未来的封棋的学生之一，在封棋未进入未来梦境之前，已经通过了猩红研究院的考核，成为了组织的成员，负责保护被暗杀的人类天才。\\[5pt]
根据文章内容，混入猩红研究院开启卧底计划的是“黑影”。在封棋与黑影的对话中，黑影透露自己来自一个由多个弱小族群组成的联盟，负责刺杀人类新生代天才等行动。同时，黑影也提到自己和他的族群并不像科技研究院或虎魄研究院那般强大，而是被迫在外打拼，只能通过暗杀等方式潜入人类社会，表明其立场和背景。\\[5pt]
根据您提供的文章，混入猩红研究院并开启卧底计划的角色是“黑影”。黑影是掌控科技研究院的潜伏者，与黑玄派系的灵能族成员有关，暗中培养力量，目的是为了对抗族地灵能族。有关黑影的信息在文章的多处被提及，其作用和动机与整个情节的发展密切相关。\\[5pt]
% 根据文本内容，猩红研究院的秘密基地由黑影控制，而黑影的身份是灵能族的叛徒。因此，可以推测是黑影混入猩红研究院开启了他的卧底计划。在猩红研究院内，黑影可能以极高的权限进行操作，确保自身的目的得以实现。\\[5pt]
\textbf{Prediction:} 黑影
% \textbf{Analysis:} The original text repeatedly mentions who infiltrates the Scarlet Research Institute and begins his undercover mission. The Map Reduce method combines all related information, leading to an inaccurate prediction. Our method, by scoring and prioritizing the most relevant information, produces an accurate answer.
\end{tcolorbox}
% \caption{Example for Zh.QA}
% \end{figure}
% \twocolumn

\newpage

\subsection{Case: Long Survey Generation}

% In the following, we present a comparison of one subsection generated by our method and AutoSurvey for the same topic. The survey generated by \ExtAgents presents a more related and more diverse range of innovative LLM applications in education, making the overall argument more persuasive and resulting in overall 68 pages compared to 58 pages from AutoSurvey.  

% \onecolumn
\begin{tcolorbox}[enhanced,
  breakable, before upper={\setlength{\parskip}{5pt}}, colback=green!10!white, colframe=green!40!black, title=A Subsection in Long Survey Generated by \ExtAgents]
\textbf{1.6 Case Studies Demonstrating LLM Impact on Education}

The integration of Large Language Models (LLMs) into educational contexts has resulted in transformative changes, showcasing their potential to enhance teaching, learning, and administrative processes. Numerous case studies illustrate the successful implementation of LLMs in various educational settings, providing insights into their effectiveness and the tangible benefits they deliver.

One notable case study focused on using LLMs to simulate student learning behaviors, where researchers leveraged LLMs to create virtual student models that replicate real learner patterns based on demographic data. This experiment involved 145 participants and revealed that the simulated results aligned closely with the actual students’ learning behaviors across diverse demographics. This application demonstrates how LLMs can enhance inclusivity in curriculum design by providing insights into how different student characteristics influence learning outcomes [40].

In the realm of personalized learning, case studies have highlighted the development of LLM-driven intelligent tutoring systems. These systems adapt to individual student needs, providing real-time support tailored to specific learning styles and paces. In one study, LLMs were implemented as personalized tutors in mathematics, demonstrating that such systems could significantly improve comprehension and engagement among learners. The ability to receive tailored guidance directly addresses students' challenges, ultimately enhancing educational efficacy [5].

Automated grading systems powered by LLMs have also gained traction in educational institutions. These systems evaluate students' assignments consistently and objectively, thereby saving educators time and minimizing biases often associated with manual grading processes. In practical studies, LLMs have shown reliable scoring that correlates well with human assessment, allowing educators to focus more on quality instruction rather than administrative tasks [41].

Interactive tools facilitated by LLMs, such as chatbots, have revolutionized student engagement models. A significant case involved an LLM-based tool providing on-demand programming assistance to students in an introductory computer science course. Over a period of 12 weeks, the tool managed more than 2,500 queries, primarily related to immediate help with assignments. The findings suggested that students who engaged frequently with the tool had higher success rates in the course, indicating that LLM-powered assistance can significantly enhance the learning experience in large classroom settings [42].

Furthermore, LLMs have been effectively employed to enhance language acquisition among learners. One study examined how situational dialogue models fine-tuned on LLMs facilitate conversational practice for language students. The models allowed for rich, simulated dialogues that mirrored authentic conversations, leading to significant improvements in fluency and confidence. Participants who engaged with these LLM-driven dialogues exhibited enhanced communication skills compared to those who did not, showcasing the crucial role of LLMs in supporting language learning [43].

In the context of automated content generation, LLMs have proven capable of producing educational resources quickly and effectively. A noteworthy implementation is the development of AutoPlan, which leverages LLMs to create interactive decision-making tasks. This intelligent system significantly increases the efficiency of educational environments centered on decision-making and planning, illustrating how LLMs can enhance problem-solving instruction by breaking down complex tasks into manageable components [44].

Moreover, LLMs have demonstrated their capability to contribute to informal learning situations through chat-based environments. Researchers have explored how learners utilize LLMs to seek answers outside traditional classroom boundaries, supporting knowledge acquisition and empowering students to take charge of their own learning. The results indicate that LLM interactions not only facilitate knowledge gain but also promote lifelong learning habits that are essential in a rapidly changing world [45].

Another compelling case involved utilizing LLMs to support health literacy in educational settings. A comprehensive assessment using various prompts with several LLMs, including ChatGPT and Google Bard, demonstrated their effectiveness in tailoring health-related content for different reading levels. The findings revealed that LLMs could produce responses at appropriate reading levels for younger audiences, thus enhancing health communication capabilities and aligning with the goal of promoting understanding among diverse learner populations [46].

A unique implementation highlights how LLMs can serve as reading companions for students. Through a study involving a user-centric assistant named Claude, researchers found tangible improvements in reading comprehension and engagement among students utilizing the AI tool compared to traditional study methods. This innovative use of LLMs showcases their potential as supplemental educational resources, helping learners tackle complex literature and academic texts effectively while fostering enhanced interaction with reading materials [47].

Moreover, the case of employing LLMs in hybrid workplace decision support exemplifies their adaptability and decision-making capabilities. In educational contexts, these systems guide students in creating effective study plans and managing their time better. By analyzing users' decision-making processes, LLMs assist learners in balancing various factors, thus enhancing their academic experience [48].

Finally, research has begun to explore the ethical implications and potential biases associated with LLM implementations in education. One study examined how LLMs were evaluated for ethical considerations in real-world educational practices, underscoring the importance of transparent LLM development processes. The findings emphasized that ethical reflection and intervention could significantly enhance educational outcomes while addressing biases, highlighting the dual need for technological innovation and ethical responsibility when deploying LLMs in educational settings [49].

These case studies collectively illustrate the transformative impact of LLMs in various educational contexts. They highlight the models' dual role as supportive tools for educators and personalized learning facilitators for students. From dialogic interactions and intelligent tutoring systems to automated content generation and tailored health communication, the versatility of LLM applications is evident. However, they also reveal the complexities and challenges of integrating AI into educational systems, necessitating a balanced approach that prioritizes ethical considerations, pedagogical integrity, and student engagement.

As the landscape of education continues to evolve, the integration of LLMs holds immense potential for enriching the learning experience. Continuous research and iterative refinements will be essential to harnessing the full capabilities of LLMs, ensuring they complement traditional educational practices while addressing the diverse needs of learners. The future of education may present unprecedented opportunities through the effective deployment of these advanced AI systems, paving the way for personalized, engaging, and impactful learning experiences that prepare students for the challenges of the 21st century.
\end{tcolorbox}
% \captionof{figure}{Long Survey Generated by ExtAgents}

\clearpage
\begin{tcolorbox}[breakable, enhanced, before upper={\setlength{\parskip}{5pt}}, colback=gray!10!white, colframe=gray!50!black, title=A Subsection in the Long Survey Generated by AutoSurvey]
\textbf{1.6 Case Studies Demonstrating LLM Impact on Education}

Large Language Models (LLMs) have taken center stage in educational innovation, demonstrating significant potential to enhance teaching and learning experiences across various contexts. This subsection presents a variety of case studies that illustrate successful deployments of LLM technologies in educational settings, showcasing their effectiveness and the tangible benefits achieved in real-world scenarios.

One notable application of LLMs is highlighted in the study titled "Future-proofing Education: A Prototype for Simulating Oral Examinations Using Large Language Models." This research explored a prototype system designed to simulate oral examinations in higher education. Educators and students evaluated this system, which aimed to automate and enhance the examination process. The outcomes showed that the prototype provided personalized feedback to students while significantly reducing the workload on educators, demonstrating the LLM's capability to streamline assessment processes and improve educational efficiency [27].

Another innovative implementation involved using LLMs to simulate student learning behavior. In the paper "Leveraging generative artificial intelligence to simulate student learning behavior," researchers revealed how they utilized LLMs to craft virtual students with distinct demographics. The findings from three experiments showcased the ability of LLMs to replicate intricate learning behaviors and experiences, unveiling correlations between course materials, engagement levels, and understanding. This simulation approach empowers educators to design curricula that adapt dynamically to diverse student needs, thus enhancing inclusivity and educational effectiveness [29].

Additionally, the integration of LLMs within role-playing scenarios has proven effective in fostering engagement and active learning among students. The study titled "Role-Playing Simulation Games using ChatGPT" illustrated how incorporating ChatGPT into role-playing simulations could enhance the quality of teaching by providing realistic environments for learners to practice skills. This approach led to increased student interest and participation, exemplifying the potential of LLMs to create immersive and interactive learning experiences [30].

LLMs have also shown promise in automating the grading process, a crucial area of concern in educational assessments. In "Three Questions Concerning the Use of Large Language Models to Facilitate Mathematics Learning," the study discusses how LLMs can provide adaptive feedback on mathematical problem-solving tasks. By assessing students' answers, LLMs can identify misconceptions and offer tailored guidance, potentially leading to better learning outcomes. Such systems emphasize the transition toward more personalized assessment methods in education [31].

Moreover, the implementation of LLMs in content generation showcases their value in educational resource development. The research titled "Prototyping the use of Large Language Models (LLMs) for adult learning content creation at scale" examined how LLMs could assist in creating quality learning materials for adult education. By leveraging a human-in-the-loop approach, the study found that LLMs could produce high-quality content quickly, marking significant advancements in Generative AI's application for education. This capacity to automate content creation not only reduces the burden on educators but also ensures resource availability and accessibility [32].

In a collaborative learning context, LLMs have been utilized to foster peer interaction and engagement. The paper titled "The Use of Multiple Conversational Agent Interlocutors in Learning" explored how LLMs could simulate conversations among various personas in an educational environment. This method assists learners in problem-solving by exposing them to different viewpoints and areas of expertise, enriching the collaborative learning experience. Such interactive dynamics can enhance comprehension and engagement among students, urging creative exploration of subject matter [33].

Implementing LLMs within existing educational frameworks also presents advantages for developing personalized learning experiences. The study "Adapting Large Language Models for Education: Foundational Capabilities, Potentials, and Challenges" provides insights into how LLMs can adapt to individual student requests, addressing diverse learning obstacles. By utilizing LLMs capable of interpreting and responding to students' needs, educational institutions can design more effective and customized learning modules, reinforcing the model's role as a facilitator in education [34].

Furthermore, the integration of LLMs into LMS (Learning Management Systems) has revealed additional benefits. The research "Identifying Critical LMS Features for Predicting At-risk Students" discussed how integrating LLMs into LMSs could analyze student interaction data to identify those at risk of underperforming. By using predictive analytics, educators can intervene proactively, thus promoting student success through personalized academic support [35].

Finally, ethical considerations, crucial when implementing LLMs in educational environments, have been addressed in the study "The teachers are confused as well: A Multiple-Stakeholder Ethics Discussion on Large Language Models in Computing Education." The authors conducted stakeholder interviews to explore the ethical challenges faced by both educators and students in using LLMs. Their findings underscore the importance of establishing guidelines and fostering digital literacy among students to ensure a balanced approach to utilizing LLMs responsibly in education [25].

Overall, these case studies collectively illustrate the transformative impact of LLMs in education. By automating assessments, enhancing content generation, enriching collaborative learning experiences, and facilitating personalized education, LLMs hold the potential to revolutionize pedagogical practices. As educational institutions continue to explore and refine LLM applications, efforts to address ethical concerns and optimizing their deployment will be critical to fully realizing their capabilities in fostering an effective and inclusive learning environment. The integration of LLMs can pave the way for innovative educational solutions and provide a solid foundation for future research and development in the field.
\end{tcolorbox}
% \captionof{figure}{Long Survey Generated by AutoSurvey}
\twocolumn

\begin{table*}[t]
    \centering
    \resizebox{0.6\linewidth}{!}{
    \begin{tabular}{lcccccc}
    \toprule
    \multirow{2}{*}{\textbf{Method}} & \multicolumn{2}{c}{\textbf{HotpotQA}} & \multicolumn{2}{c}{\textbf{En.QA}} & \multicolumn{2}{c}{\textbf{Zh.QA}} \\ \cmidrule(lr){2-3} \cmidrule(lr){4-5} \cmidrule(lr){6-7}
     &  Helmet &  Input & Helmet &  Input &  Helmet &  Input \\
    \midrule\midrule
    \multicolumn{7}{c}{\textit{DeepSeek-R1-Distill-Llama-8B}} \\
    \textcolor{gray}{Direct Input} & \textcolor{gray}{1.56}  & \textcolor{gray}{32k} &  \textcolor{gray}{0.69} &  \textcolor{gray}{32k} &  \textcolor{gray}{0.66} &  \textcolor{gray}{32k} \\
    \midrule
    \multicolumn{7}{c}{\textit{gpt-4o-mini-2024-07-18}} \\
    Direct Input &  1.83 &  128k &  1.41 &  128k &  1.04 &  128k \\
    DRAG &  1.53 &  128k & \\
    IterDRAG &  1.70 &  128k &   \\
    LLM$\times$MapReduce &   &  &  1.12 &  256k &  1.04 &  128k \\
    \textbf{ExtAgents} (Ours) &  \textbf{1.71} &  \textbf{1024k} &  \textbf{1.20} &  \textbf{1024k} &  \textbf{1.10} &  \textbf{256k} \\
    \midrule
    \multicolumn{7}{c}{\textit{Llama-3.1-8B-Instruct}} \\
    Direct Input &  0.96 &  128k &  0.93 &  128k &  0.89 &  128k \\
    DRAG &  1.20 &  32k &  \\
    IterDRAG &  1.14 &  32k &  \\
    Chain of Agents &   &   &  0.51  &  32k &  0.57  &  32k \\
    LLM$\times$MapReduce &  &   &  0.78 &  256k &  0.79 &  256k \\
    \textbf{ExtAgents} (Ours) &  \textbf{1.38} &  \textbf{1024k} &  \textbf{1.09} &  \textbf{1024k} &  \textbf{0.85} &  \textbf{256k} \\
    \midrule
    \multicolumn{7}{c}{\textit{gpt-4o-2024-08-06}} \\
    \ExtAgents ($N=1$) &  1.73 &  128k &  \\
    \textbf{\ExtAgents} &  \textbf{1.86} &  \textbf{1024k} &  \\
    \bottomrule
    \end{tabular}
    }
    \caption{Performance on Multi-Hop QA tasks in Helmet correctness scores with the optimal setting and the corresponding input length (\#tokens). The settings are the same as Table~\ref{tab:qa}.}
    \label{tab:qa-helmet}
\end{table*}

\section{Implementation Details}
\label{app:implementation}

\subsection{Experiment Details}

All datasets and models are used under their individual licenses without intention to violate any terms. 

For evaluation metrics, 
we found existing principles~\citep{wang2024autosurvey,wang2025llmtimesmapreducev2entropydrivenconvolutionaltesttime} for LLM-based judges have very low discriminative power on generated surveys, and thus we use the prompting template from \citet{liu2025inferencetimescalinggeneralistreward} to judge the result against baselines. 
We also include the Helmet correctness score~\citep{yen2025helmet} and additional quantitative measurement (e.g., \#citations, citation density, duplication rate, etc.) as supplementary metrics for QA and long generation tasks, respectively. 
We control the whole input length for each methods in the range of $\{8k, 16k, 32k, 64k, 128k, 256k, 512k, 1024k\}$ tokens with the maximum context window $128k$, and also control the chunk sizes of each agent for LLM-based multi-agent methods (Figure~\ref{fig:scaling-exp}). Specifically, for a minority of samples with contexts longer than $1024k$, we input all contexts for the $1024k$ setting.

%TODO: fix
We utilized several large language models in our experiments. Closed-source models included gpt-4o-mini-2024-07-18 and gpt-4o-2024-08-06, accessed via API. For these models, the sampling temperature was set to 0. Open-source models employed were Llama-3.1-8B-Instruct and Llama-3.2-3B-Instruct. These models were deployed on four NVIDIA A100 80GB GPUs, and the sampling temperature was set to 0.1. The maximum input context length was 128,000 tokens for the closed-source models and 131,092 tokens for the open-source models.  
We test these methods with the optimal configuration of input lengths and chunk sizes (Table~\ref{tab:qa}). For stable reproduction, we report the median results of three runs. For HotpotQA, we use BM25 retriever~\citep{10.1561/1500000019}, and for AutoSurvey, we use the original retrieval method. 

Baseline implementations are adjusted slightly for each task. For En.QA and Zh.QA test sets, the direct input method is implemented using the official InfiniteBench repository \citep{zhang-etal-2024-bench}. The LLM$\times$MapReduce is re-implemented by us to align with our settings with \ExtAgents. For HotpotQA, we re-implement the DRAG and IterDRAG methods due to the lack of official code, utilizing the prompts provided in the appendix of \citet{yue2025inferencescalinglongcontextretrieval}. For the survey generation task, the AutoSurvey baseline was adapted from the official implementation~\citep{wang2024autosurvey}, with the reflection and refinement removed from the original pipeline. For the metrics, the citation density is calculated as the number of citations divided by the number of thousand tokens in the generated survey. The duplication rate is calculated as the number of duplicate citations divided by the total number of citations. The LLM-as-a-Judge method on long survey generation is implemented with gpt-4.1-mini-2025-04-14~\citep{openai2025gpt45}, with temperature set to 1.  

For \ExtAgents, in En.QA and Zh.QA tests, information extracted from different chunks is ranked based on scores rated by Seek Agents, based on the task query. For the HotpotQA test, the information is ranked based on the retrieval priority of chunks. The chunk exclusion mechanism is disabled in long-document QA tasks. For AutoSurvey task, we set the number of chunks of input retrieved papers to 4, and thus assigning 4 Seeking Agents to input papers. Other hyper-parameters are set to the default values. The overall process for our methods involves a maximum of $T=5$ synchronization timesteps. We only employ the knowledge accumulation strategy at $t=1$. This strategy processes information incrementally, starting with the top 1, then top 2, top 4, top 8, and finally all information ($S=5$), following a power-of-two sequence. The whole process halts once the answer is obtained.

\subsection{Additional Experimental Results}
\label{app:add-result}

\begin{table}[t]
    \centering
    \resizebox{0.9\linewidth}{!}{
    \begin{tabular}{lcccc}
    \toprule
    \multirow{2}{*}{\textbf{Method}} & \multicolumn{2}{c}{\textbf{En.QA}} & \multicolumn{2}{c}{\textbf{Zh.QA}} \\ \cmidrule(lr){2-3} \cmidrule(lr){4-5} 
     & F1 &  Input &  F1 &  Input \\
    \midrule\midrule
    \multicolumn{5}{c}{\textit{DeepSeek-R1-Distill-Llama-8B}} \\
    \textcolor{gray}{Direct Input} & \textcolor{gray}{.104} &  \textcolor{gray}{32k} &  \textcolor{gray}{.144} &  \textcolor{gray}{32k} \\
    \midrule
    \multicolumn{5}{c}{\textit{gpt-4o-mini-2024-07-18}} \\
    Direct Input &  .189 &  128k &  .206 &  128k \\
    LLM$\times$MapReduce &   .385 &  128k &  .443 &  128k \\
    \textbf{\ExtAgents} (Ours) &   \textbf{.421} &  \textbf{1024k} &  \textbf{.491} &  \textbf{1024k} \\
    \midrule
    \multicolumn{5}{c}{\textit{Llama-3.1-8B-Instruct}} \\
    Direct Input &   .267 &  128k &  .316 &  128k \\
    LLM$\times$MapReduce &  .287 &  256k &  .347 &  128k \\
    \textbf{\ExtAgents} (Ours) &   \textbf{.322} &  \textbf{512k} &  \textbf{.352} &  \textbf{256k} \\
    \bottomrule
    \end{tabular}
    }
    \caption{Performance on $\infty$Bench with the optimal setting and the corresponding input length (\#tokens). Other settings are the same as Table~\ref{tab:qa}.}
    \label{tab:qa-origin}
\end{table}

\paragraph{Helmet Correctness Scores on Multi-Hop QA}
We provide corresponding Helmet correctness scores~\citep{yen2025helmet} on multi-hop QA tasks complementary to Table~\ref{tab:qa} in Table~\ref{tab:qa-helmet}. F1 scores may misjudge the performance of LLMs, especially when the response is long and the answer is not concise enough. In our experiments, we observe that the trend of Helmet correctness scores are consistent with the F1 scores, indicating that the performance of \ExtAgents is robust and reliable.

\paragraph{Results on the Original $\infty$Bench}
We provide the result on the original $\infty$Bench in Table~\ref{tab:qa-origin}. We observe the same trend as in $\infty$Bench+ in Table~\ref{tab:qa}, that \ExtAgents achieves the highest performance with the longest input contexts. And other methods, including inference-time scaling with long CoT, fail to utilize longer contexts beyond the context window and reach inferior results.

% \paragraph{Results on Open-Source LLMs}
% Due to space constraints, we only show the results on gpt-4o-mini-2024-07-18 in the main text. The results on Llama-3.1-8B-Instruct are shown in Figure~\ref{fig:llama3}. We observe that \ExtAgents consistently outperforms the baseline methods across all input lengths, demonstrating its effectiveness in scaling external knowledge input.

\begin{figure*}[t]
    \centering
    % \begin{subfigure}[b]{0.235\linewidth}
    % \includegraphics[width=\linewidth]{./figs/Figures_PDF/RAG_gpt-4o-mini/f1_v3_16k.pdf}
    % \end{subfigure}
    % \begin{subfigure}[b]{0.235\linewidth}
    % \includegraphics[width=\linewidth]{./figs/Figures_PDF/RAG_gpt-4o-mini/f1_v3_64k.pdf}
    % \end{subfigure}
    % \begin{subfigure}[b]{0.235\linewidth}
    % \includegraphics[width=\linewidth]{./figs/Figures_PDF/RAG_gpt-4o-mini/drag_and_iterdrag.pdf}
    % \end{subfigure}
    % \noindent\parbox{\textwidth}{\rule{\textwidth}{0.1pt}}
    % \begin{subfigure}[b]{0.235\linewidth}
    % \includegraphics[width=\linewidth]{./figs/Figures_PDF_enhanced/f1_llama_v19_enhanced_16k.pdf}
    % \end{subfigure}
    % \begin{subfigure}[b]{0.235\linewidth}
    % \includegraphics[width=\linewidth]{./figs/Figures_PDF_enhanced/f1_llama_v19_enhanced_64k.pdf}
    % \end{subfigure}
    % \begin{subfigure}[b]{0.235\linewidth}
    % \includegraphics[width=\linewidth]{./figs/Figures_PDF_enhanced/f1_llama_enhanced_all.pdf}
    % \end{subfigure}
    % \noindent\parbox{\textwidth}{\rule{\textwidth}{0.1pt}}
    % \begin{subfigure}[b]{0.235\linewidth}
    % \includegraphics[width=\linewidth]{./figs/Figures_PDF_enhanced/f1_v19_zh_enhanced_16k.pdf}
    % \end{subfigure}
    % \begin{subfigure}[b]{0.235\linewidth}
    % \includegraphics[width=\linewidth]{./figs/Figures_PDF_enhanced/f1_v19_zh_enhanced_64k.pdf}
    % \end{subfigure}
    % \begin{subfigure}[b]{0.235\linewidth}
    % \includegraphics[width=\linewidth]{./figs/Figures_PDF_enhanced/f1_zh_enhanced_all.pdf}
    % \end{subfigure}
    \includegraphics[width=0.7\linewidth]{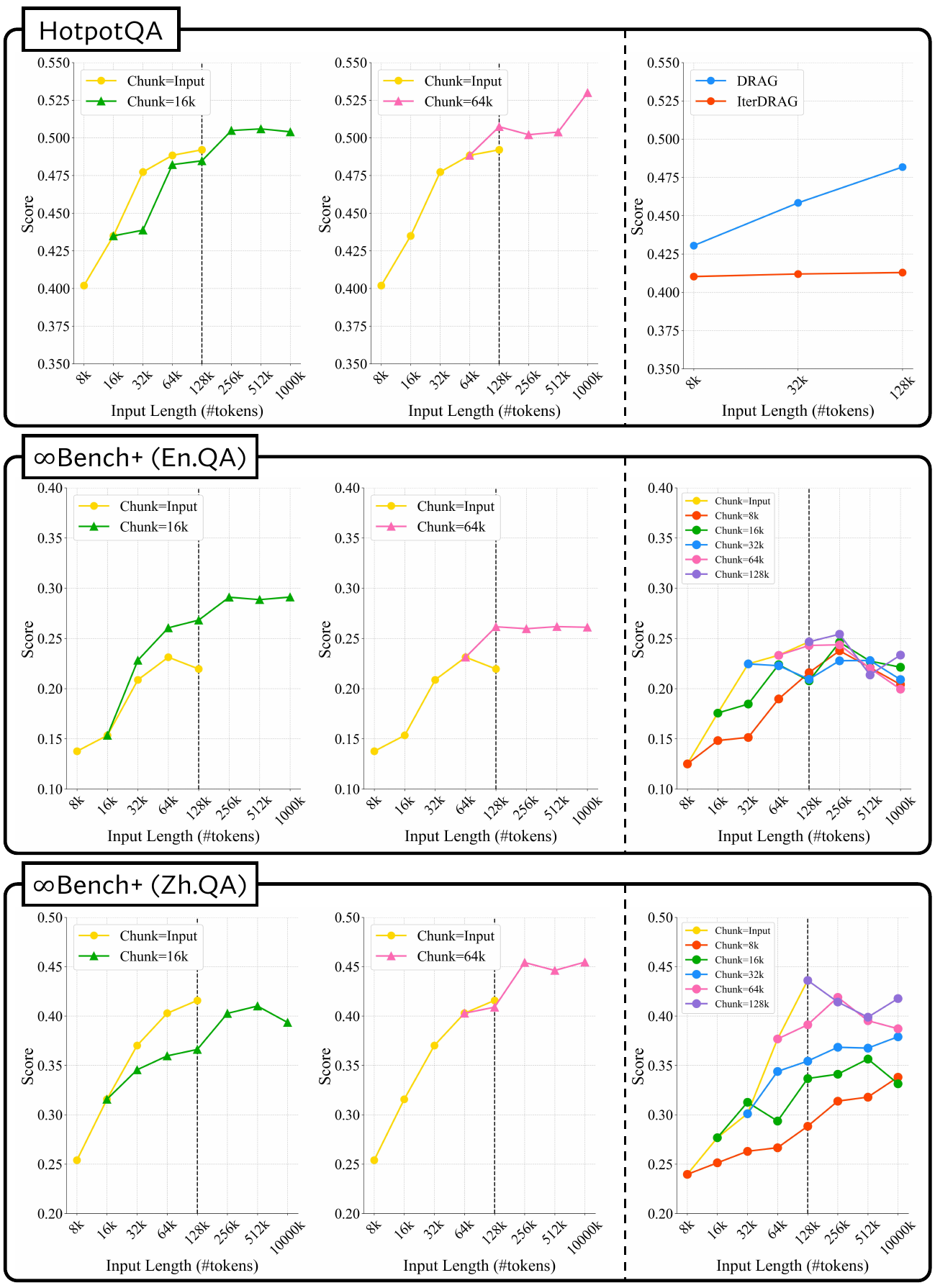}
    \caption{Detailed experimental results of scaling external knowledge input on multi-hop QA tasks, complementary to Figure~\ref{fig:scaling-exp} with the same subfigure arrangement.} 
    \label{fig:scaling-exp-detail}
\end{figure*}

\begin{figure*}[t]
    \centering
    \includegraphics[width=\linewidth]{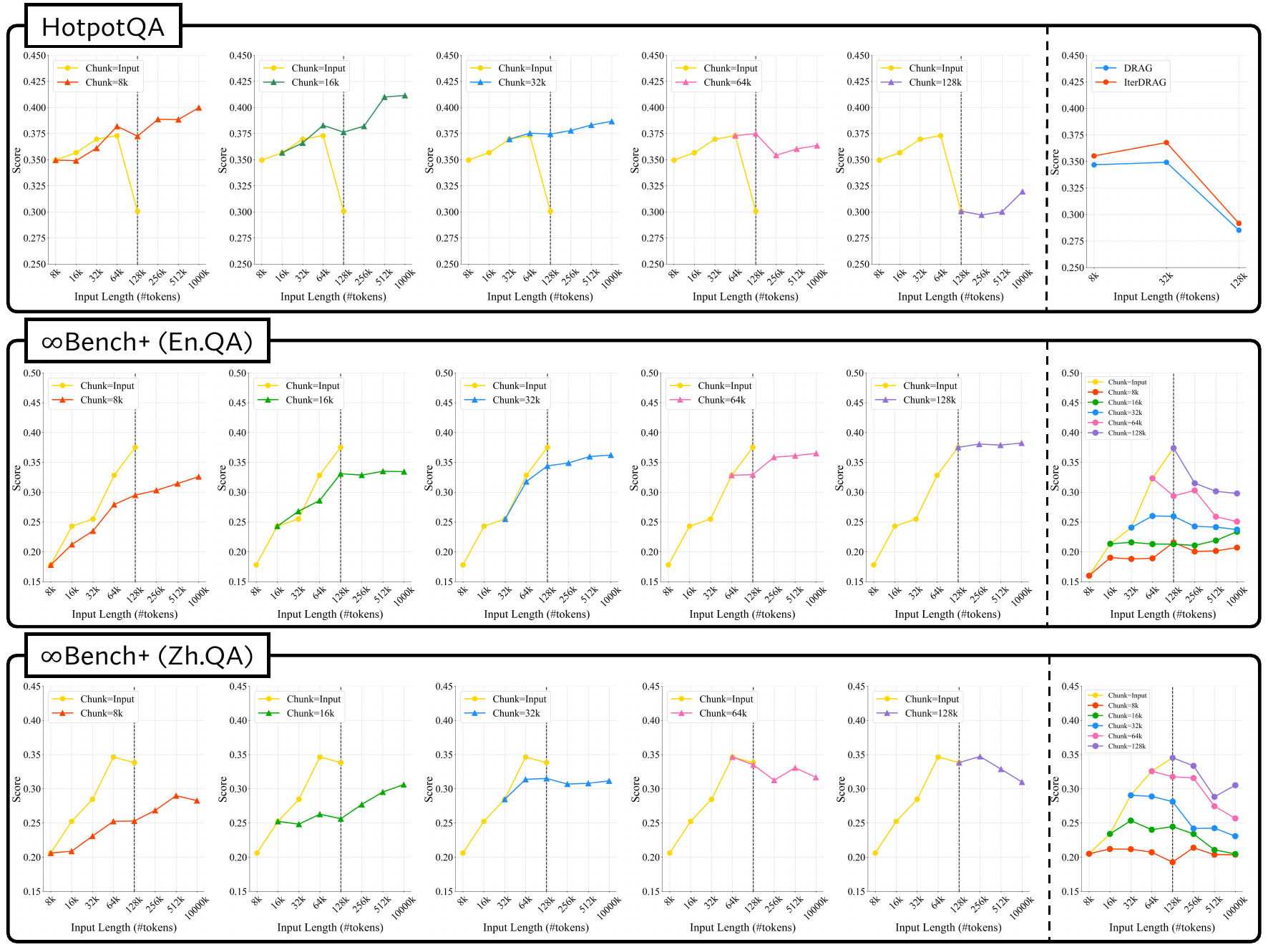}
    \caption{Experiment of scaling external knowledge input on multi-hop QA tasks. We plot gpt-4o-mini results for En.QA and Llama-3.1-8B-Instruct results for other tasks. Other arrangement of subfigures is the same as Figure~\ref{fig:scaling-exp}. }
    \label{fig:scaling-exp-llama}
\end{figure*}

\paragraph{Detailed Results on Multi-Hop QA}
We provide the detailed results on multi-hop QA tasks in Figure~\ref{fig:scaling-exp-detail} and Figure~\ref{fig:scaling-exp-llama}. The results are consistent with the main findings in Figure~\ref{fig:scaling-exp} with enriched results. 

% \begin{wrapfigure}[14]{r}{0.25\linewidth}
%     \centering
%     \vspace{-3em}
%     \includegraphics[width=\linewidth]{./figs/Figures_PDF/RAG_gpt-4o/f1_4o_v3_128k.pdf}
%     \caption{Results of \ExtAgents with gpt-4o-2024-08-06 on HotpotQA benchmark.}
%     \label{fig:stronger-llm}
% \end{wrapfigure}

\begin{figure*}[b]
    \centering
    % \begin{subfigure}[b]{0.22\textwidth}
    %     \includegraphics[width=\textwidth]{./figs/Figures_PDF/RAG_llama3.2_3B/f1_llama3b_v3_8k.pdf}
    %     % \caption{1}
    % \end{subfigure}%
    % \hfill
    % \begin{subfigure}[b]{0.22\textwidth}
    %     \includegraphics[width=\textwidth]{./figs/Figures_PDF/RAG_llama3.2_3B/f1_llama3b_v3_32k.pdf}
    %     % \caption{2}
    % \end{subfigure}%
    % \hfill
    % \begin{subfigure}[b]{0.22\textwidth}
    %     \includegraphics[width=\textwidth]{./figs/Figures_PDF/RAG_llama3.2_3B/f1_llama3b_v3_128k.pdf}
    %     % \caption{3}
    % \end{subfigure}
    % \hfill
    % \begin{subfigure}[b]{0.22\textwidth}
    %     \includegraphics[width=\textwidth]{./figs/Figures_PDF/RAG_llama3.2_3B/drag_and_iterdrag_llama3b.pdf}
    %     % \caption{3}
    % \end{subfigure}
    \includegraphics[width=\linewidth]{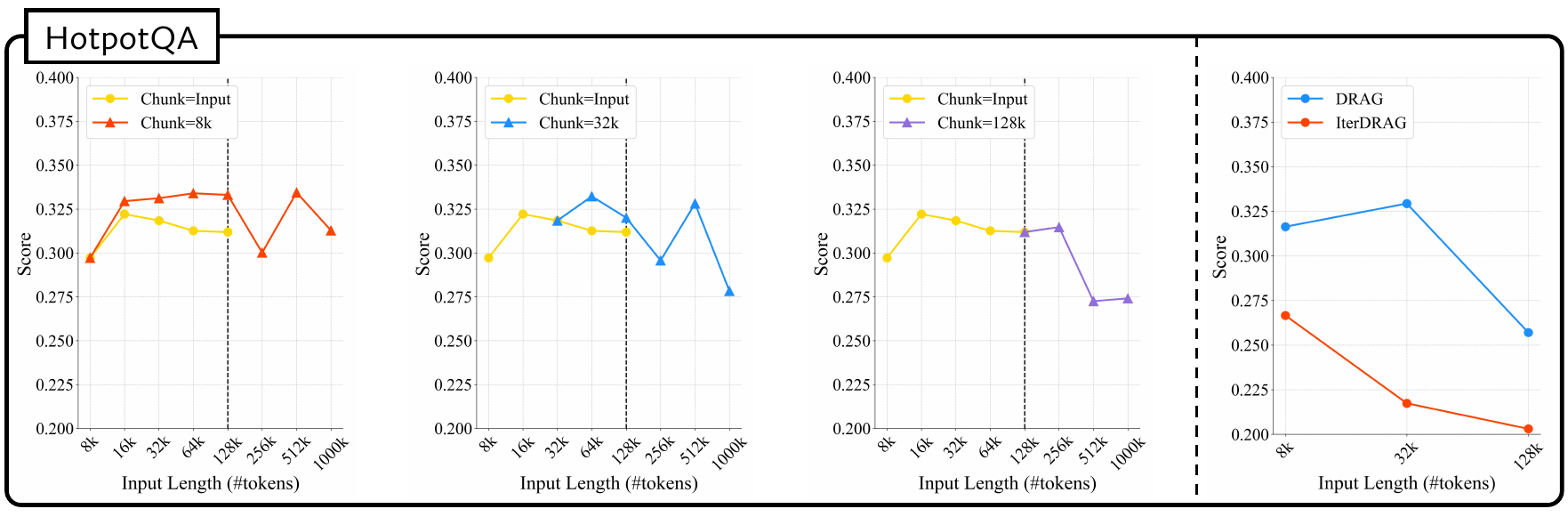}
    \caption{Results of \ExtAgents with Llama-3.2-3B-Instruct on HotpotQA benchmark.}
    \label{fig:weaker-llm}
\end{figure*}

\paragraph{Results on Weaker \& Stronger LLMs}
We also test the performance of \ExtAgents on a weaker LLM, Llama-3.2-3B-Instruct, besides a stronger LLM, gpt-4o-2024-08-06, on HotpotQA benchmark. The results are shown in Figure~\ref{fig:weaker-llm} and Figure~\ref{fig:stronger-llm}, respectively. We observe that \ExtAgents achieves consistent performance improvements over scaled external knowledge input, and the performance gap is larger on the stronger model, potentially due to the better collaboration capability of stronger LLMs.

% \paragraph{Detailed Costs Analysis\quad}
\begin{table}[t]
    \centering
    \resizebox{0.9\linewidth}{!}{
    \begin{tabular}{lccc}
    \toprule
    \textbf{Method} & \textbf{HotpotQA} & \textbf{En.QA} & \textbf{Zh.QA} \\
    \midrule\midrule
    DRAG & 0.019 &  \\
    IterDRAG & 0.048 &  \\
    LLM$\boldsymbol{\times}$MapReduce &  & 0.022 & 0.021 \\
    \midrule
    \ExtAgents (Ours) & 0.021 & 0.025 & 0.029 \\
    \bottomrule
    \end{tabular}
    }
    \caption{Average costs (\$) of \ExtAgents and baseline methods on gpt-4o-mini-2024-07-18.}
    \label{tab:cost-detail}
\end{table}

\paragraph{Detailed Costs Analysis}
We demonstrate the average costs of \ExtAgents and baseline methods on gpt-4o-mini-2024-07-18 in Table~\ref{tab:cost-detail}. 
The costs are calculated based on the average number of tokens in the input and output, where the cost of 1M token input is \$0.15 and 1M token output is \$0.60. The extra cost of \ExtAgents is due to the global knowledge synchronization, which introduces larger bandwidth and according costs. 

\begin{table*}[t]
    \centering
    \resizebox{0.68\linewidth}{!}{
    \begin{tabular}{lcccccccc}
    \toprule
    \multirow{2}{*}{\textbf{Method}} & \multicolumn{4}{c}{\textbf{En.QA}} & \multicolumn{4}{c}{\textbf{Zh.QA}} \\ \cmidrule(lr){2-5} \cmidrule(lr){6-9} 
     & F1 & Helmet &  Input & Chunk &  F1 &  Helmet &  Input & Chunk \\
    \midrule\midrule
    \multicolumn{9}{c}{\textit{DeepSeek-R1-Distill-Llama-8B}} \\
    \textcolor{gray}{Direct Input} & \textcolor{gray}{.097} & \textcolor{gray}{0.69}  &  \textcolor{gray}{32k} &  \textcolor{gray}{-} &  \textcolor{gray}{.143} &  \textcolor{gray}{0.66} &  \textcolor{gray}{32k} &  \textcolor{gray}{-} \\
    \midrule
    \multicolumn{9}{c}{\textit{gpt-4o-mini-2024-07-18}} \\
    \textbf{\ExtAgents} (Ours) &   \textbf{.382} & \textbf{1.20} & \textbf{1024k} & 128k & \textbf{.482} & \textbf{1.10} &  \textbf{1024k} &  128k \\
    \midrule
    \multicolumn{9}{c}{\textit{Llama-3.1-8B-Instruct}} \\
    Direct Input &   .237 & 0.93 &  128k & - & .315 & 0.89 &  128k & - \\
    \multirow{5}{*}{Chain of Agents} &   .168 & 0.51 &  32k & 32k & .246 & 0.57 &  32k & 32k \\
    &   .075 & 0.27 &  128k & 32k & .123 & 0.32 &  128k & 32k \\
    &   .027 & 0.12 &  128k & 128k & .070 & 0.21 &  128k & 128k \\
    &   .073 & 0.27 &  1024k & 32k & .158 & 0.34 &  1024k & 32k \\
    &   .068 & 0.21 &  1024k & 128k & .064 & 0.20 &  1024k & 128k \\
    LLM$\times$MapReduce &  .254 & 0.78 &  256k & 128k & .345 & 0.79 &  128k & 128k \\
    \textbf{\ExtAgents} (Ours) &   \textbf{.291}  & \textbf{1.09}  &  \textbf{1024k} & 16k &  \textbf{.347} & \textbf{0.85} & \textbf{256k} & 128k \\
    \bottomrule
    \end{tabular}
    }
    \caption{Performance of Chain of Agents on $\infty$Bench+ in different configurations of chunk sizes and input lengths (\#tokens). Results of other methods are picked from Table~\ref{tab:qa}. F1 and Helmet correctness scores are reported.}
    \label{tab:coa-detail}
    \vspace{1em}
    \resizebox{0.82\linewidth}{!}{
    \begin{tabular}{lccc}
    \toprule
    \textbf{Forced Answer Correctness} & \textbf{Model Judged "Can Answer"} & \textbf{Model Judged "Cannot Answer"} & \textbf{Total} \\ \midrule
    \textbf{Answer is Correct} & 149 (True Positives) & 21 (False Positives) & 170 \\
    \textbf{Answer is Incorrect} & 130 (False Negatives) & 71 (True Negatives) & 201 \\
    \bottomrule
    \end{tabular}
    }
    \caption{Sample distribution (\#samples) of the answerability check in \ExtAgents, comparing model judgement against accuracy when forced to answer, demonstrating effective safeguarding and confident hallucination.}\label{tab:answerability}
\end{table*}

\paragraph{Scaling External Knowledge Input with Chain of Agents}
We test the scaling performance of Chain of Agents under several configurations of input lengths and chunk sizes. Results are shown in Table~\ref{tab:coa-detail}. Alike \citet{zhou2024llmtimesmapreducesimplifiedlongsequenceprocessing}, we also observes inferior scaling behaviors in Chain of Agents compared to LLM$\times$MapReduce and \ExtAgents, which also aligns to our theoretical analysis of the bottleneck of knowledge synchronization (Chain of Agents has a smaller bandwidth value of constant 2, according to Table~\ref{tab:existing-methods}). Specifically, Chain of Agents failed to achieve steady performance improvement when adding more input length with either the same or different chunk sizes. Generally, it degrades performance when inputting from 32k to 1024k, not able to handle overwhelming external knowledge. 
Also, Chain of Agents could not achieve better than directly inputting 128k tokens into the LLM, let alone \ExtAgents with 1024k external knowledge input.

\paragraph{Robustness Analysis of the Answerability Check}
\ExtAgents halts on Reasoning Agent deciding the query is answerable or hitting the maximum synchronization timestep $T$, and then outputting the answer (Section~\ref{sec:extagents}). 
To examine the robustness of the answerability check, we conducted a quantitative analysis of false positive answers versus appropriate refusals. We isolated this mechanism by removing the Knowledge-Accumulating Reasoning (KAR) strategy and providing exactly 4 chunks of information to the Reasoning Agent, whose role is to identify answerability and refuse insufficient queries. Using gpt-4o-mini, we evaluated 371 samples in HotpotQA to observe how the model's self-judged answerability aligns with its actual ability to answer correctly when forced. As shown in Table~\ref{tab:answerability}, results reveal: (1) \textbf{Effective Safeguarding} (False Positives): Only in 21/170 cases that model answers correctly, the model judged that it could not answer. This indicates the answerability check does not prevent correct answers from being outputted. 
(2) \textbf{Confident Hallucinations} (False Negatives): The model exhibited overconfidence in 130/201 cases, believing it had sufficient information to answer but ultimately providing an incorrect response. This highlights the inherent limitations of current LLMs regarding the self-assessment of their own knowledge bounds. It could be alleviated with stronger models, and we indeed observe stronger models have better end-to-end performance (Section~\ref{sec:results-anal}).

\begin{table}[t]
\centering
\resizebox{\linewidth}{!}{
\begin{tabular}{lcccc}
\toprule
\textbf{Input Length (\#Tokens)} & \textbf{T=1} & \textbf{T=2} & \textbf{T=5 (Default)} & \textbf{T=10} \\
\midrule
\phantom{0}32k & .456 & .454 & .477 & .480 \\
128k & .485 & .496 & .500 & .507 \\
512k & .493 & .494 & .516 & .521 \\
\bottomrule
\end{tabular}
}
\caption{Impact of the maximum synchronization timesteps ($T$) in \ExtAgents with gpt-4o-mini on HotpotQA (F1 Score). The chunk size is fixed to 32k tokens.}
\label{tab:sensitivity_t}
\end{table}

\begin{table}[t]
\centering
\resizebox{0.85\linewidth}{!}{
\begin{tabular}{lc}
\toprule
\textbf{Scheduling Strategy} & \textbf{F1 Score} \\
\midrule
Base-2 Logarithmic (Seq: $1, 2, 4, 8\dots$) & .516 \\
Base-4 Logarithmic (Seq: $1, 4, 16\dots$) & .509 \\
Proportional Fractional ($k=\lfloor N/2 \rfloor$) & .468 \\
Exhaustive Accumulation ($k=N$) & .506 \\
Small Window Constraint ($k=4$) & .439 \\
\bottomrule
\end{tabular}
}
\caption{Impact of reasoning iteration scheduling ($S$) in \ExtAgents with gpt-4o-mini on HotpotQA. The input length is 512k tokens and the chunk size is 32k.}
\label{tab:sensitivity_s}
\end{table}

\paragraph{Detailed Ablations on Hyper-Parameters}
To evaluate the hyperparameter sensitivity of \ExtAgents, we investigated the influence of the maximum synchronization timesteps ($T$) and the reasoning iteration scheduling strategy ($S$) on HotpotQA using gpt-4o-mini with a 32k-token chunk size. As detailed in Table \ref{tab:sensitivity_t}, expanding the maximum synchronization timesteps consistently enhances F1 scores across various input context lengths, demonstrating the framework's ability to effectively leverage extended inference-time compute for complex queries. However, the performance improvements exhibit diminishing marginal utility beyond $T=5$, validating our default configuration as an optimal equilibrium between reasoning accuracy and computational overhead. Furthermore, Table \ref{tab:sensitivity_s} illustrates the efficacy of different context accumulation schedules. Our default base-2 logarithmic scheduling (Seq: $1, 2, 4, 8\dots$) achieves the highest score. Static strategies—such as a small window ($k=4$) or a proportional window ($k=\lfloor N/2 \rfloor$)—deprive the Reasoning Agent of sufficient global context. Conversely, exhaustive accumulation degrades performance by reintroducing the information overload bottleneck characteristic of long-context processing, even with ranked orders.

\onecolumn

\twocolumn

\onecolumn

% \twocolumn

% \onecolumn

\clearpage
\subsection{Prompt Templates}~\label{app:prompts}

% We enumerate the prompt templates used in experiments in the order of tasks. 

% \begin{figure}[H]
\begin{tcolorbox}[enhanced,
  breakable,colframe=blue!75!black, colback=blue!10!white, title=Prompt Templates for HotpotQA]
\textbf{Knowledge synchronization: First iteration}\\[5pt]
We are working on long-text question answering, and you are responsible for one chunk. Read the following chunk and extract as much information as possible related to the question. Ensure your extracted information provides clear context and is logically complete. If no information, just output "NO INFORMATION".\\[5pt]
Your chunk:\\
\{Retrieved documents\}\\[5pt]
Question:\{question\}\\[5pt]
\textbf{Knowledge synchronization: Other iterations}\\[5pt]
We are working on long-text question answering, and you are responsible for one chunk. This is the \{iteration\} round of Q\&A. And we have the previously extracted information from all chunks in the previous round. Based on the previously extracted information and question, extract new information from the chunk. Do not repeat the previously extracted information. If no new information, just output "NO INFORMATION".\\[5pt]
Your chunk:\\
\{Retrieved documents\}\\[5pt]
Previously extracted information:\\
\{Previously extracted information\}\\[5pt]
Question: \{question\}\\[5pt]
\textbf{Knowledge-accumulating reasoning: No need to terminate}\\[5pt]
We have the following extracted information from different chunks of the text:\\[5pt]
\{Extracted information\}\\[5pt]
Based on the extracted information, decide whether you can confidently answer the question. If you can, combine and reduce this information into a final answer, as short as possible, word or phrase. If you cannot, just output "NO ANSWER".\\[5pt]
Question: \{question\}\\[5pt]
\textbf{Knowledge-accumulating reasoning: Need to terminate}\\[5pt]
We have the following extracted information from different chunks of the text:\\[5pt]
\{Extracted information\}\\[5pt]
Based on the extracted information, combine and reduce this information into a final answer, as short as possible, word or phrase.\\[5pt]
Question: \{question\}
\end{tcolorbox}
% \caption{Example prompt for RAG}
% \end{figure}

\clearpage
% \begin{figure}[H]
\begin{tcolorbox}[enhanced,
  breakable,colframe=blue!75!black, colback=blue!10!white, title=Prompt Templates for En.QA in $\infty$Bench+]
\textbf{Knowledge synchronization: First iteration}\\[5pt]
Read the following article and extract as much information as possible related to the question.\\[5pt]
\{Context\}\\[5pt]
Question: \{question\}\\[5pt]
\textbf{Knowledge synchronization: Other iterations}\\[5pt]
We are working on long-text question answering, and you are responsible for one chunk. This is the \{iteration\} round of Q\&A. And we have the previously extracted information from all chunks in the previous round. Based on the previously extracted information and question, extract new information from the chunk. Do not repeat the previously extracted information.\\[5pt]
Your chunk:\\
\{Context\}\\[5pt]
Previously extracted information:\\
\{Extracted information\}\\[5pt]
Question: \{question\}\\[5pt]
\textbf{Knowledge synchronization: Ranking information}\\[5pt]
Based on the extracted information and question, provide a score (0-100) for how useful the extracted information is for answering this question.\\[5pt]
Extracted information: \{extracted information\}\\[5pt]
Question: \{question\}\\[5pt]
Please follow this format:\\[5pt]
Score: (0-100)\\[5pt]
\textbf{Knowledge-accumulating reasoning: No need to terminate}\\[5pt]
We have the following extracted information from different chunks of the text:\\[5pt]
\{Extracted information\}\\[5pt]
Based on the extracted information, decide whether you can confidently answer the question. If you can, combine and reduce this information into a final answer, as short as possible, word or phrase. If you cannot, just output "NO ANSWER".\\[5pt]
Question: \{question\}\\[5pt]
\textbf{Knowledge-accumulating reasoning: Need to terminate}\\[5pt]
We have the following extracted information from different chunks of the text:\\[5pt]
\{Extracted information\}\\[5pt]
Based on the extracted information, combine and reduce this information into a final answer, as short as possible, word or phrase.\\[5pt]
Question: \{question\}
\end{tcolorbox}
% \caption{Example prompt for En.QA}
% \end{figure}

\clearpage
% \begin{figure}[H]
\begin{tcolorbox}[enhanced,
  breakable,colframe=blue!75!black, colback=blue!10!white, title=Prompt Template for Zh.QA in $\infty$Bench+]
\textbf{Knowledge synchronization: First iteration}\\[5pt]
请阅读以下文章并尽可能提取与问题相关的信息。\\[5pt]
\{Context\}\\[5pt]
问题：\{question\}\\[5pt]
\textbf{Knowledge synchronization: Other iterations}\\[5pt]
我们正在进行长文本问答任务，你负责处理其中一个文本块。这是第\{iteration\}轮问答。我们在之前几轮已经对所有文本块中提取了信息。请基于先前提取的信息和问题，从当前文本块中提取新信息。不要重复已提取的信息。\\[5pt]
你的文本块：\\
\{Context\}\\[5pt]
先前提取的信息：\\
\{Extracted information\}\\[5pt]
问题：\{question\}\\[5pt]
\textbf{Knowledge synchronization: Ranking information}\\[5pt]
根据提取的信息和问题，给出一个分数（0-100），评估提取的信息对回答该问题的有用程度。\\[5pt]
提取的信息：\{extracted information\}\\[5pt]
问题：\{question\}\\[5pt]
请遵循以下格式：\\[5pt]
Score: (0-100)\\[5pt]
\textbf{Knowledge-accumulating reasoning: No need to terminate}\\[5pt]
我们有以下从不同文本块中提取的信息：\\[5pt]
\{Extracted information\}\\[5pt]
根据提取的信息，请判断是否能确定地回答该问题。如果能，将这些信息合并并简化为最终答案。请尽量简短地回答，只使用一个或多个词语。如果不能，直接输出"NO ANSWER".\\[5pt]
问题：\{question\}\\[5pt]
\textbf{Knowledge-accumulating reasoning: Need to terminate}\\[5pt]
我们有以下从不同文本块中提取的信息：\\[5pt]
\{Extracted information\}\\[5pt]
根据提取的信息，将这些信息合并并简化为最终答案。请尽量简短地回答，只使用一个或多个词语。\\[5pt]
问题：\{question\}
\end{tcolorbox}
% \caption{Example prompt for Zh.QA}
% \end{figure}
\twocolumn

\end{CJK*}

\end{document}